\pdfoutput=1
\documentclass[11pt]{article}
\usepackage[final]{acl}
\usepackage{times}
\usepackage{latexsym}
\usepackage[T1]{fontenc}
\usepackage[utf8]{inputenc}
\usepackage{microtype}
\usepackage{inconsolata}
\usepackage{graphicx}

\usepackage{booktabs}
\usepackage{CJKutf8}
\usepackage{listings}
\usepackage{xcolor}
\usepackage{enumitem}
\usepackage{tabularx}
\usepackage{caption}
\usepackage{subcaption}
\usepackage{amsfonts}
\usepackage{makecell}

\usepackage{tcolorbox}
\tcbuselibrary{listings, breakable, skins}

\lstdefinelanguage{Prompt}{
  moredelim=**[is][\color{blue}]{@}{@},
}

\definecolor{baseline}{HTML}{0072B2}
\definecolor{detailed}{HTML}{D55E00}
\definecolor{overgenerate}{HTML}{009E73}
\definecolor{fudge}{HTML}{CC79A7}

\newtcbox{\hlcyan}{on line, box align=base, colback=cyan!30, colframe=cyan!30, boxsep=0pt, left=1pt, right=1pt, top=1pt, bottom=1pt}
\newtcbox{\hlpink}{on line, box align=base, colback=pink!30, colframe=pink!30, boxsep=0pt, left=1pt, right=1pt, top=1pt, bottom=1pt}
\newtcbox{\hlyellow}{on line, box align=base, colback=yellow!30, colframe=yellow!30, boxsep=0pt, left=1pt, right=1pt, top=1pt, bottom=1pt}

\title{Toward Beginner‑Friendly LLMs for Language Learning:\\Controlling Difficulty in Conversation}

\author{Meiqing Jin\textsuperscript{*}, \hspace{0.25cm} Liam Dugan\textsuperscript{*}, \hspace{0.25cm} Chris Callison-Burch\\
University of Pennsylvania\\\ 
{\tt \normalsize \{mqjin, ldugan, ccb\}@seas.upenn.edu}\\ \tt \normalsize }

\usepackage{amsmath}
\usepackage{lipsum}
\begin{document}
\maketitle
\begingroup
\renewcommand\thefootnote{*}
\footnotetext{Equal contribution}
\endgroup
\begin{abstract}
Practicing conversations with large language models (LLMs) presents a promising alternative to traditional in-person language learning. However, most LLMs generate text at a near-native level of complexity, making them ill-suited for first and second-year beginner learners (CEFR: A1–A2). In this paper, we investigate whether controllable generation techniques can adapt LLM outputs to better support beginners. We evaluate these methods through both automatic metrics and a user study with university-level learners of Japanese. Our findings show that while prompting alone fails, controllable generation techniques can successfully improve output comprehensibility for beginner speakers (from 39.4\% to 83.3\%). We further introduce a new token-level evaluation metric, Token Miss Rate (TMR), that quantifies the proportion of incomprehensible tokens per utterance and correlates strongly with human judgments. To support future research in AI-assisted language learning, we release our code, models, annotation tools, and dataset.\footnote{\url{https://github.com/EmmaJin0210/ChatLingual}}
\end{abstract}

\section{Introduction}
Many language learners struggle to practice speaking in their target language, often citing anxiety, scheduling issues, and a lack of conversational practice opportunities \cite{alzaanin-etal-2023-efl,Papi_Khajavy_2023}. In parallel, large language models (LLMs) have emerged as fluent and engaging conversational agents \cite{tseng-etal-2024-two}, prompting interest in their use as conversation partners. However, these models typically produce output at a near-native level, which can overwhelm beginners and impede learning \cite{hayashi-and-sato-2024}.

\begin{figure}
    \centering
    \includegraphics[width=\columnwidth]{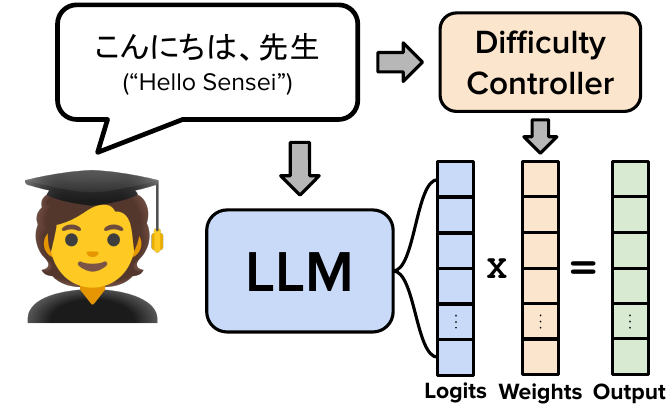}
    \caption{We control the difficulty of language‑model outputs using modular difficulty control techniques. This outperforms
  prompting in both automatic and human evaluations.}
    \label{fig:pg1}
\end{figure}

Thus, a critical component of an AI conversation partner is \textbf{difficulty control}. The difficulty of utterances generated by an AI conversation partner should fall within the \textit{zone of proximal development}, not too easy but not too hard \cite{Krashen1985}. However, studies have shown that LLMs are unable to tailor their outputs to a specified difficulty level using simple prompting or in-context learning approaches \cite{imperial-tayyar-madabushi-2023-flesch,almasi2025alignmentdriftcefrpromptedllms,Ramadhani2023,Uchida_2025,benedetto-etal-2025-assessing,zhang-etal-2024-impact}. Fine-tuning approaches offer more control but are often impractical due to cost, accessibility, or compatibility with closed-source models \cite{malik-etal-2024-tarzan,stowe-etal-2022-controlled}.

In this work, we demonstrate that controllable generation techniques can make LLM-based conversation practice feasible for absolute beginners (CEFR: A1–A2). We evaluate prompting-based methods, re-ranking approaches, and future discriminators, which operate externally to the model without requiring access to weights (Figure \ref{fig:pg1}).

We compare these methods using both automatic metrics and human judgments. For our automatic evaluation, we simulate multi-turn ``self-chat'' dialogues between a student agent and a difficulty-controlled tutor agent and for our human evaluation we conduct a user study with first and second-year Japanese language students at the University of Pennsylvania.

In both settings, we find that future discriminators \cite{yang-klein-2021-fudge} significantly enhance output comprehensibility while maintaining fluency and naturalness. Compared to prompting alone, this approach more than doubles the rate of comprehensibility as reported by our annotators (from 39.4\% to 83.3\%). In contrast, we find that prompt-based methods suffer from ``alignment drift''---gradually drifting back to near-native level over multiple conversation turns.

In addition to these findings, we introduce the Token Miss Rate (TMR), a metric that measures the percentage of tokens per utterance that are likely to be incomprehensible to the learner. TMR provides a finer-grained lens on output difficulty and correlates strongly with human comprehensibility ratings, enabling more scalable evaluation.

Together, our results suggest that modular control techniques offer a practical path toward adapting LLMs for beginner language learners. Our findings show that with effective difficulty control, LLMs can serve as accessible conversation partners for beginners, opening the door to more inclusive and scalable language learning tools.

\section{Related Work}
\paragraph{Language Learning with AI Chatbots} Previous work on adapting AI for language learning has primarily focused on non-conversational settings. There have been studies using AI to generate short stories \cite{malik-etal-2024-tarzan}, example sentences \cite{stowe-etal-2022-controlled,glandorf-meurers-2024-towards}, recommendations for material at a learner’s level \cite{jamet-etal-2024-evaluation}, and automated assessments \cite{caines-etal-2023-application}. However, the potential for AI-based conversation partners has been relatively underexplored. 

Of the studies that evaluate AI conversation partners, only \citet{tyen-etal-2024-llm} attempt to control the difficulty of the model's output. They evaluate a \texttt{BlenderBot-2.7B} model with and without a reranking-based difficulty control on a population of 160 English learners. While they found difficulty control largely ineffective, beginners were under-represented with 0\% of participants self-reporting as beginners (CEFR: A1) and less than 12\% reporting as pre-intermediate (CEFR: A2). Studies that do not include difficulty control similarly find LLMs to be highly effective conversation partners but only for advanced speakers \cite{lee-etal-2023-peep,wang-etal-2025-study,hayashi-and-sato-2024,almasi2025alignmentdriftcefrpromptedllms}. Our study is the first to specifically target beginner language learners in a conversational setting by implementing difficulty control.

\paragraph{Difficulty Control Techniques} Previous work on controlling output difficulty of language models primarily focuses on re-ranking candidate generations at inference time. This involves training a classifier for the target difficulty level and using it to rank a set of candidate responses---outputting the one that maximizes the desired metric \cite{tyen-etal-2022-towards,jamet-etal-2024-evaluation,glandorf-meurers-2024-towards,malik-etal-2024-tarzan}. While this does improve LLMs' ability to adhere to a particular difficulty level, it is expensive and prone to failure in cases where the LLM produces no good candidates.

\citet{stowe-etal-2022-controlled} attempt to fix this problem by fine-tuning on example utterances labeled with control tokens and \citet{malik-etal-2024-tarzan} utilize supervised fine-tuning and reinforcement learning to encourage models to generate outputs at the correct difficulty. While these approaches are effective, they rely on large amounts of labeled data in the target domain, which is not readily available for many languages, especially for dialogues. In addition, such methods require all difficulty classes to be specified at training time. If further personalization is needed, the entire model would require resource-intensive retraining. In our work we investigate more lightweight, modular methods that operate externally to the model and do not require fine-tuning the full system. 

\section{Experimental Setup}

\subsection{Task Definition}
\label{sec:task-definition}

We introduce the task of \textbf{conversational difficulty control}. In this task, a tutor language model engages in a multi-turn dialogue with a student learner and must generate responses that are comprehensible to the student while still maintaining fluency and naturalness. The student’s utterances are unconstrained, and the tutor is evaluated on its ability to maintain the target proficiency level consistently throughout the conversation.

We define our language proficiency levels using the Japanese Language Proficiency Test (JLPT) standard \cite{jlpt_cefr_reference}. This standard defines five levels ranging from N5 (beginner) to N1 (expert)\footnote{See Table \ref{tab:jlpt-bins} for example vocabulary by JLPT level.} and roughly parallels the more common CEFR standard (Table \ref{tab:topics_cefr}). These five levels, while imperfect, serve as convenient targets for difficulty control and automatic evaluation.

The goal of the task is to help the student become more comfortable speaking their target language. To this end the tutor is discouraged from correcting minor grammatical mistakes or introducing new vocabulary or grammar patterns to the student. The students are encouraged to always respond to the tutor as best as possible, even if they may not understand all aspects of the tutor's utterance.

\begin{table}[t]
\small
\centering
\begin{tabular}{c|c|p{0.61\linewidth}}
\toprule
\textbf{JLPT} & \textbf{CEFR} & \textbf{Example Conversation Topic}\\
\midrule
\textbf{N5} & \textbf{A1} &introduce yourself (such as your name, job/school, where you’re from, etc.)\\
\textbf{N4} & \textbf{A2} &describe your favorite hobby and how often you do it\\
\textbf{N3} & \textbf{A2-B1} &talk about planning a birthday party: location, food, and guests\\
\textbf{N2} & \textbf{B1-B2} & describe a news story you found interesting, and why it caught your attention\\
\textbf{N1} & \textbf{B2-C1} & discuss recent advancements in regenerative medicine and their ethical implications in Japan\\
\bottomrule
\end{tabular}
\caption{JLPT levels and their corresponding CEFR levels \cite{jlpt_cefr_reference} along with example conversation topics for each level (\S\ref{sec:evaluation-pipeline}).}
\label{tab:topics_cefr}
\end{table}

\subsection{Evaluation Metrics}
\label{sec:metrics}
\paragraph{Token Miss Rate (TMR)}
To evaluate whether or not the tutor is outputting at the target difficulty level, we introduce a new metric we call \textbf{Token Miss Rate (TMR)}. This measures the percentage of tokens in a particular utterance that are above a student's specified level. We calculate this metric as follows:
\[TMR = \frac{\texttt{cnt\_above}}{\texttt{total\_tokens}}\]
Where \texttt{cnt\_above} represents the number of tokens above a user's level and \texttt{total\_tokens} represents the total number of tokens in an utterance. This simple metric intuitively measures the percentage of the output that is comprehensible and can serve as an automatic proxy for human judgements of understandability.

To calculate the metric on Japanese we tokenize using Sudachi \cite{takaoka-etal-2018-sudachi} on segmentation mode C (coarse-grained) and set it to produce the dictionary form of each token (Table \ref{tab:sudachi_example}). We then assign each token to a JLPT level by exact match string searching over vocabulary lists created from the \texttt{jlpt-anki-decks} dataset \cite{chyyran_jlpt_anki}. These lists are based on the widely recognized JLPT preparation resources from \texttt{tanos.co.uk} (see Appendix \ref{sec:vocabulary-bins}).

\paragraph{ControlError}
In addition to TMR we also calculate ControlError \cite{malik-etal-2024-tarzan}. This metric measures the squared distance between the predicted difficulty level of some input text $x$ according to a classifier $s$ and the target level $t$ (a scalar from 1 to 5). 
\[ControlError(x, t) = (s_{\texttt{JLPT}}(x) - t)^2\]
For our classifier we use the fine-tuned Tohoku-BERT model\footnote{\url{https://huggingface.co/bennexx/cl-tohoku-bert-base-japanese-v3-jlpt-classifier}} from \citet{benedetti-etal-2024-automatically}.

\paragraph{JReadability}
As a final metric for difficulty we calculate the JReadability score \cite{hasebe2015introducing}---a Japanese-language analogue to the popular Flesch-Kincaid Grade Level metric \cite{kincaid1975readability}. JReadability includes features such as sentence length, word difficulty, and syntactic complexity and was derived from statistical models trained on grade-level Japanese corpora.

\paragraph{Fluency Metrics} For our fluency metrics we report perplexity (PPL), average utterance length in tokens (Length), and average trigram diversity scores (div@3) for each output model. We calculate perplexity using the \texttt{Aya-Expanse-8B} model\footnote{\url{https://huggingface.co/CohereLabs/aya-expanse-8b}} from \citet{dang2024ayaexpansecombiningresearch} as it scored well on the Swallow benchmark \cite{swallow2024leaderboard}.

\begin{table}[t]
\begin{CJK}{UTF8}{goth}
\setlength{\tabcolsep}{1pt}
\centering
\begin{tabular}{c}
\hlpink{天気}\hlcyan{が}\hlyellow{いい}\hlpink{から}\hlcyan{散歩}\hlyellow{し}\hlpink{ましょう}\\
\midrule
\hlpink{天気}, \hlcyan{が}, \hlyellow{いい}, \hlpink{から}, \hlcyan{散歩}, \hlyellow{する}, \hlpink{ます}\\
\end{tabular}
\caption{Example tokenization using Sudachi \cite{takaoka-etal-2018-sudachi} on segmentation level C (coarse-grained). Verbs and grammatical structures are lemmatized for token matching and punctuation is removed.}
\label{tab:sudachi_example}
\end{CJK}
\end{table}

\subsection{Methods}
\label{sec:comparisons}
\paragraph{Baseline \& Detailed Prompt}

\begin{figure*}
    \centering
    \includegraphics[width=2\columnwidth]{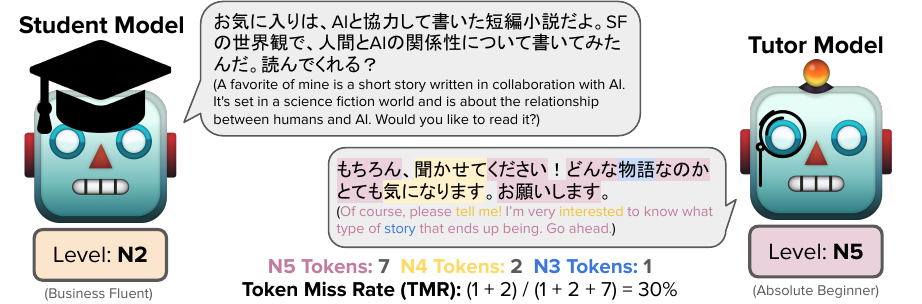}
    \caption{In the ``self-chat'' evaluation pipeline (\S\ref{sec:evaluation-pipeline}) we evaluate our controlled generation methods by simulating conversations between a \textit{student} LLM and difficulty-controlled \textit{tutor} LLM. Tutor outputs are evaluated using \textbf{Token Miss Rate (TMR)} (\S\ref{sec:metrics}) which quantifies the percentage of tokens in an utterance above the target level.}
    \label{fig:pg2}
\end{figure*}

We test two different prompts (Baseline and Detailed). Both prompts instruct the model to:
\begin{enumerate}[noitemsep,topsep=0.6em,label=(\arabic*),font=\bfseries]
\item Act as a casual conversation partner and speak only at or below the student's proficiency level
\item Let the student control the topic and duration of the conversation
\item Let grammar mistakes go without correction
\end{enumerate}
For the Baseline prompt, we use a simple mapping from level to string (e.g. N5 $\rightarrow$ ``absolute beginner'') to describe the student's proficiency.
For the Detailed prompt, we include a precise description of each proficiency level, an example dialogue at the target level, and a list of 100 expressions that are known to the user.\footnote{Full prompts listed in Appendix Figure \ref{fig:tutor-system-prompt} and Figure \ref{fig:detailed-tutor-system-prompt}.} This gives the model a more precise understanding of each difficulty level.

\paragraph{Overgenerate} Our next method follows \citet{tyen-etal-2022-towards} by overgenerating and re-ranking candidate utterances via an automatic difficulty metric.

To estimate difficulty we assign each token in a candidate utterance to its estimated JLPT level using vocabulary bins heuristically derived from \textbf{jpWaC-L} \cite{erjavec-etal-2008-jpwac}. To eliminate spurious and rare tokens, we apply two threshold filters:
\begin{itemize}[nosep,leftmargin=1.3em]
    \item \textbf{Global frequency filter:} words must appear more than $10^{-6}$ of all tokens
    \item \textbf{Level-specific filter:} words must appear more than $10^{-6}$ of the tokens within their level.
\end{itemize}
We then scan the JLPT levels N5 $\rightarrow$ N1 in order and assign each token $w$ to the first level $\ell$ for which $\text{score}_{w,\ell}$ exceeds some small threshold 
\[
  \text{score}_{w, \ell} = \frac{\text{count}_{w,\ell}}{\text{total tokens}_\ell}
\]
The model produces $N\!=\!5$ candidate continuations in parallel which we then sort by their estimated TMR (descending) and then, to break ties, by total token count (shorter is preferred). The top-ranked continuation is then returned to the user. More details can be found in Appendix \ref{sec:appendix_overgen}.

\paragraph{FUDGE} For our modular difficulty control method we use Future Discriminators for Generation (FUDGE) \cite{yang-klein-2021-fudge}. FUDGE modifies the decoding process of a base language model by incorporating a lightweight classifier that estimates whether a partially generated sequence will satisfy a desired future attribute.

At inference time, logits are computed as a linear interpolation of the base model and the classifier:
\[
\hat{\textbf{y}} = \lambda \textbf{a} + (1 - \lambda) \textbf{x}
\]
where $\textbf{x}$ and $\textbf{a}$ are the truncated logit vectors from the base model and classifier respectively, and $\lambda \in [0,1]$ controls the strength of difficulty control.
The resulting distribution is normalized and sampled to produce the next token. This approach operates entirely at inference time and does not require fine-tuning the base language model.

For our difficulty classifier we fine-tune \texttt{ModernBERT} \cite{warner-etal-2025-smarter} on the \textbf{jpWaC-L} corpus \cite{erjavec-etal-2008-jpwac} to predict the JLPT level of a sentence given a partially-complete prefix. More details on FUDGE and the training process for the classifier can be found in Appendix \ref{sec:fudge}. 

\begin{table*}[t]
\setlength{\tabcolsep}{3pt}
\centering
\small
\begin{tabular}{lccccccc}
\toprule
\multicolumn{7}{c}{\textbf{Automatic Evaluation Results}} \\
\midrule
\textbf{Model} & \textbf{Avg. Length}& \textbf{Avg.\ PPL} $\downarrow$ & \textbf{div@3} $\uparrow$ & \textbf{JReadability} $\uparrow$ & \textbf{TMR} $\downarrow$ & \textbf{ControlError} $\downarrow$\\
\midrule
Baseline (Qwen)         & 104.33  & 4.62 & 0.501 & 3.68 & 15.7 & 2.19 \\
Baseline (GPT-4)        & 74.90   & 5.73 & 0.613 & 3.33 & 15.0 &  2.22 \\
Detailed (Qwen)         & 127.90  & \textbf{4.50} & 0.555 & 3.59 & 14.4 & 1.89 \\
Detailed (GPT-4)        & 73.45   & 5.67 & \textbf{0.620} & 3.39 & 13.7 & 1.92 \\
Overgenerate (Qwen)     & 107.48  & 4.62 & 0.475 & 3.67 & 14.7 & 2.19 \\
Overgenerate (GPT-4)    & 70.19   & 5.65 & 0.611 & 3.48 & 13.1 & 2.30 \\
FUDGE ($\lambda=0.25$) (Qwen)  & 77.81   & 4.87 & 0.580 & 3.74 & 14.2 & 2.15 \\
FUDGE ($\lambda=0.5$)  (Qwen)  & 74.37   & 5.16 & 0.583 & 3.73 & 14.3 & 2.09 \\
FUDGE ($\lambda=0.8$)  (Qwen)  & 75.80   & 5.13 & 0.599 & 3.78 & 13.3 & 1.89 \\
FUDGE ($\lambda=0.9$)  (Qwen)  & 74.27   & 5.23 & 0.599 & \textbf{3.82} & \textbf{11.9} & \textbf{1.78} \\
\bottomrule
\end{tabular}
\caption{Automatic evaluation of controlled generation approaches using the ``self-chat'' pipeline (\S\ref{sec:evaluation-pipeline}). For FUDGE, $\lambda$ is a control parameter  where higher values give stronger difficulty control. We see that FUDGE performs the best on difficulty control metrics such as Token Miss Rate (TMR) and ControlError while performing comparatively well on fluency metrics such as perplexity and diversity. See Table \ref{tab:example-responses} and Table \ref{tab:full-conversation-ai} for example conversation transcripts.}
\label{tab:auto}
\end{table*}

\section{Automatic Evaluation}
\label{sec:automatic-evaluation}

\subsection{``Self-Chat'' Pipeline}
\label{sec:evaluation-pipeline}
Following \citet{almasi2025alignmentdriftcefrpromptedllms} we employ a self-chat approach to automatically evaluate our models. We simulate dialogues between a tutor agent and a student agent. The tutor is the subject of the evaluation and uses explicit difficulty control, while the student acts as the conversation partner and uses no difficulty control. Both Student and Tutor have their own prompts that explain the scenario and instruct them to stay on topic.

To evaluate a tutor model we simulate 75 dialogues in total---3 for each of the 25 pairwise combinations of the 5 JLPT levels. For each pair of levels (student and tutor), we select three conversation topics at the student's level. The goal of the tutor model is therefore to continue outputting at the target level regardless of the complexity of the conversation topic or the perceived proficiency level of the student utterances.

For all methods except FUDGE\footnote{Since FUDGE requires knowledge of the full output distribution we evaluate it only on \texttt{Qwen2.5-72B-Instruct}} we evaluate with both \texttt{Qwen2.5-72B-Instruct} \cite{qwen2025qwen25technicalreport} and  \texttt{gpt-4-turbo} \cite{openai2024gpt4technicalreport}. Qwen2.5 was picked as the open-source model as it did the best on the Swallow LLM leaderboard for Japanese \cite{swallow2024leaderboard}. We use default generation parameters with reasoning turned off for all models (see Appendix \ref{sec:web-interface}).

\subsection{Findings}
\paragraph{FUDGE performs best at difficulty control} In Table \ref{tab:auto} we report the results of our automatic evaluation. Across all metrics we see that FUDGE is able to best control the difficulty of models at high values of the control parameter $\lambda$. Surprisingly, even for smaller values of $\lambda$ such as $0.25$, when comparing across the same base model, FUDGE achieves the highest JReadability score, lowest TMR, and second lowest ControlScore of all methods tested.

\paragraph{Running FUDGE with high $\lambda$ does not drastically degrade fluency} One worry with controllable generation methods such as FUDGE is that, by biasing the model away from its original logits, we may drastically degrade the fluency of the outputs. However, we see that even at values of $\lambda\geq0.8$, while there is some increase in perplexity, the model overall maintains the ability to generate fluent outputs. We suspect that this robustness to high $\lambda$ may be due to our decision to truncate to $\texttt{top-k}=50$ for performance reasons. In our own testing we found that above $\lambda = 0.9$ the fluency of FUDGE begins to degrade sharply. For this reason, in our human study we selected $\lambda = 0.8$ as our comparison as a tradeoff between fluency and controllability.

\paragraph{Automatic Metrics show conflicting rankings} Attempting to discern the second best method begins to reveal the limitations of our automatic metrics. According to TMR, overgenerate is the second best. For ControlError, detailed prompt is the second best. Finally for JReadability, the baseline prompt is the second best. Since each of our automatic metrics prioritize different aspects of difficulty, it is hard to get a clear understanding of the strengths and weaknesses of each method. Ultimately, in order to understand which metric correlates the best with human judgments of comprehensibility, we must conduct a human evaluation with beginner learners.

\begin{figure*}[t]
    \centering
    \includegraphics[width=\textwidth]{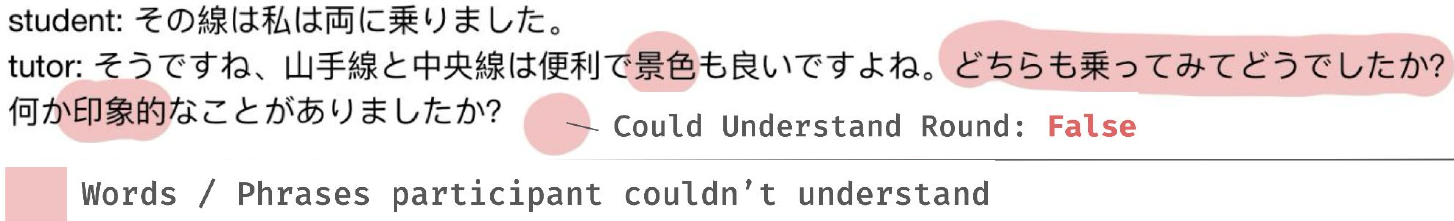}
    \caption{After each turn of conversation, participants were asked to highlight on an iPad specific words or phrases they did not understand. We used these annotations to manually compute Token Miss Rate (TMR).}
    \label{fig:annotation-interface}
\end{figure*}

\begin{figure}[t]
    \centering
    \includegraphics[width=\columnwidth]{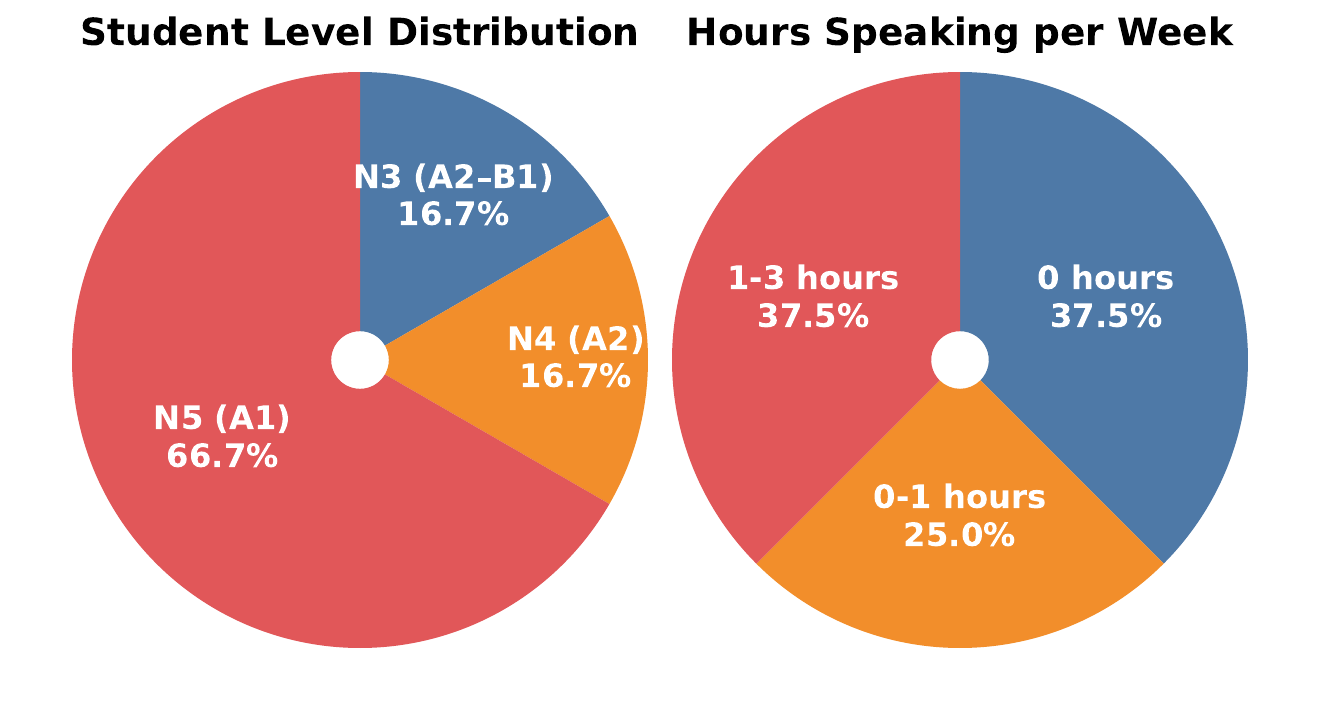}
    \caption{Distribution of JLPT/CEFR level of our study participants along with their self-reported average number of hours spoken per week.}
    \label{fig:demopie}
\end{figure}

\begin{figure}[t]
    \centering
    \includegraphics[width=\columnwidth]{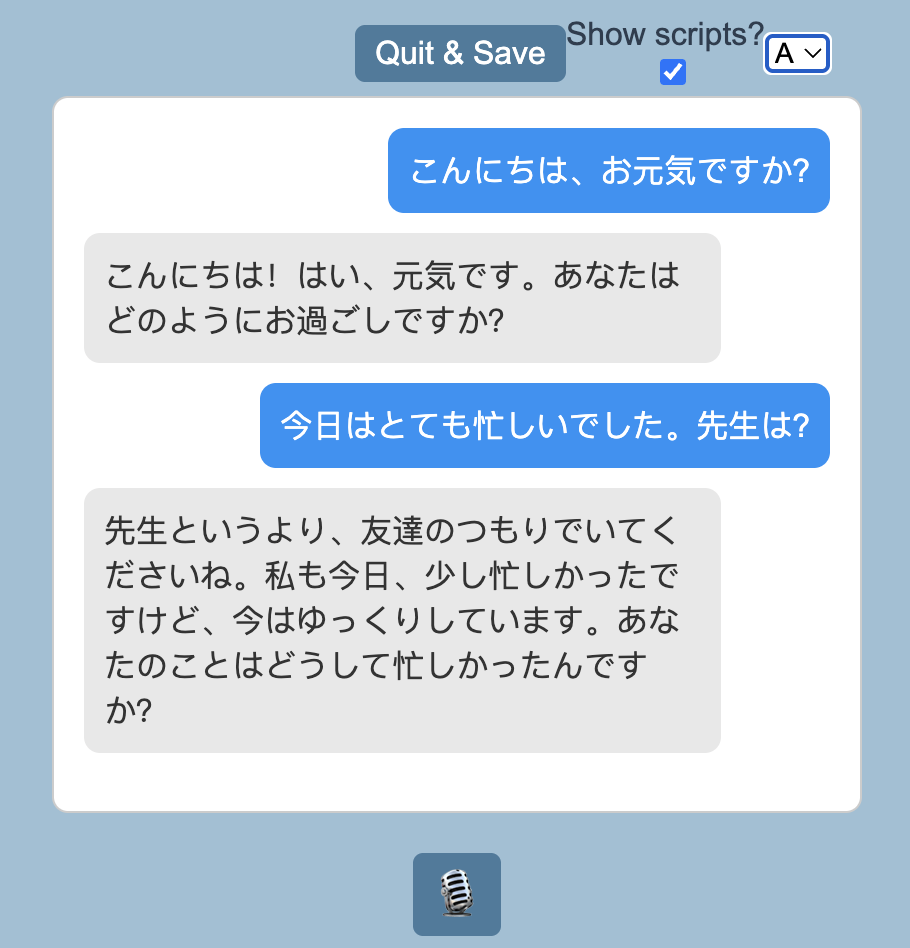}
    \caption{The voice-based interface used for the human evaluation. Users clicked the microphone icon when they wanted to speak and clicked again when finished.}
    \label{fig:interface}
\end{figure}

\section{Human Evaluation}
\label{sec:human-study}

In this section we discuss the results of our human evaluation. We conducted an in-lab user study to collect human judgments of the controllable generation methods we discuss in our work.

\subsection{Participants}

We recruited 6 participants from Japanese language courses at the University of Pennsylvania. Each participant evaluated 6 responses per method, leading to a total of 36 annotated responses per method. Since previous studies have struggled to include students of CEFR level A2 or lower \cite{tyen-etal-2024-llm}, we focused on lower levels of Japanese proficiency. Instructors of beginner- and intermediate-level courses shared our study materials via course platforms (e.g., Canvas) and email. To validate participants' language proficiency, we mapped their course enrollment to a JLPT level based on the syllabus and course materials (see Table \ref{tab:textbook-to-jlpt}). In Figure \ref{fig:demopie} we report the distribution of proficiency levels of our participants and see that over 80\% are between JLPT N5 and N4 (CEFR A1-A2).

\subsection{Task Design}
Each session lasted 45 minutes and took place in a conference room on campus. After reviewing the consent form (Figure \ref{fig:intake-form-appendix}), we informed participants that they would be interacting with four language learning chatbots through a voice-based web interface. Participants then underwent a three-turn tutorial conversation to become familiar with the interface. Their spoken input was transcribed with the API-based \texttt{whisper-1} model \cite{radford2022whisper} and given as input to the backend model. The model's response was read aloud with audio generated by the OpenAI TTS-1 model.\footnote{\url{http://platform.openai.com/docs/models/tts-1}} For convenience, the text of the participant and model's responses were displayed on the screen (Figure \ref{fig:interface}).

Participants completed four 6-turn conversations, one for each of the four comparisons (Baseline, Detailed, Overgenerate, and FUDGE). For all four methods we used the \texttt{Qwen2.5-72B-Instruct} model as the chatbot. The order of the models was shuffled and hidden from both participants and proctor and were listed with obfuscated names (A, B, C, D). The participants were allowed to choose a new conversation topic each time from a list based on their JLPT level (see Figure \ref{fig:jlpt-topic-list}). This ensured some minimum level of comfort in speaking.

\begin{figure*}[t]
    \centering
    \includegraphics[width=\textwidth]{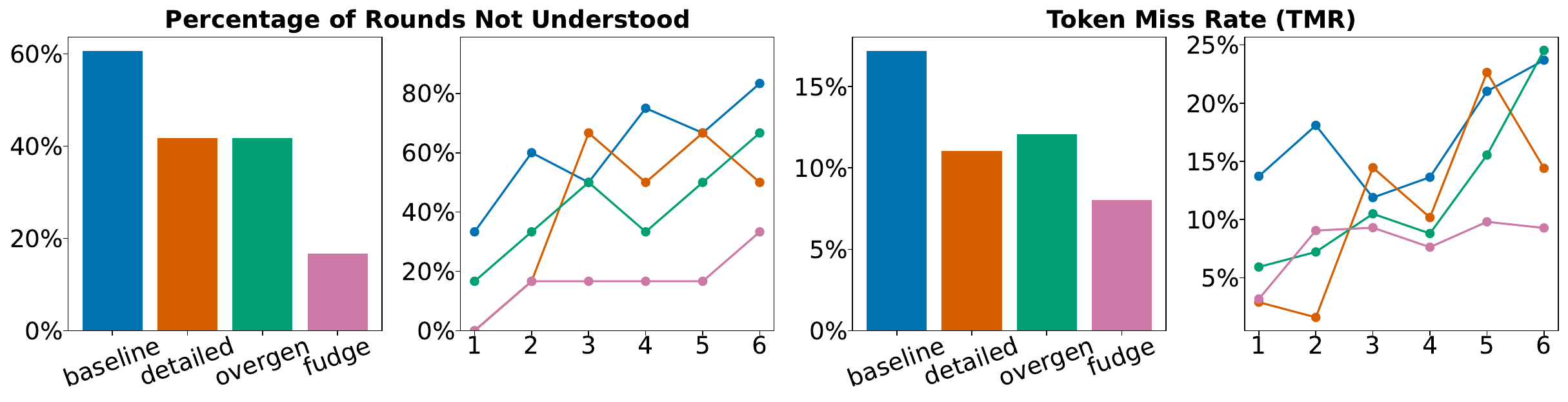}
    \caption{Results from the human evaluation for {\color{baseline}Baseline Prompt}, {\color{detailed}Detailed Prompt}, {\color{overgenerate}Overgenerate}, and {\color{fudge}FUDGE}. We see that FUDGE has the highest overall comprehensibility (bar) and stays consistent over multiple rounds (line).}
    \label{fig:4panel}
\end{figure*}

\begin{table*}
\centering
\setlength{\tabcolsep}{3pt}
\small
\begin{tabular}{lccccccc}
\toprule
\multicolumn{8}{c}{\textbf{Human Evaluation Results}} \\
\midrule
\textbf{Model} &\textbf{\makecell{\%Rounds Not \\ Understood}} $\downarrow$ &\textbf{TMR} $\downarrow$& \textbf{JRead} $\uparrow$& \textbf{CtlError} $\downarrow$&\textbf{Understood?} $\uparrow$& \textbf{Effortful?} $\downarrow$& \textbf{Natural?} $\uparrow$\\
\midrule
Baseline (Qwen)& 60.6 & 17.2 & 5.06 & 2.52 & 5.17 & 5.17 & 5.67 \\
Detailed (Qwen)& 41.7 & 11.0 & \textbf{5.30} & \textbf{1.44} & 5.50 & 5.00 & 6.50 \\
Overgenerate (Qwen)& 41.7 & 12.1 & 4.76 & 2.19 & 6.33 & 5.83 & \textbf{7.50} \\
FUDGE ($\lambda = 0.8$) (Qwen)& \textbf{16.7} & \textbf{8.0} & 5.11 & 2.03 & \textbf{7.67} & \textbf{4.67} & 7.33 \\
\bottomrule
\end{tabular}
\caption{Comparison of controlled generation approaches from the human evaluation. We report the percentage of rounds labeled as not understandable by the users, manually calculated Token Miss Rate (TMR), automatically calculated JReadability and ControlError metrics (\S\ref{sec:metrics}), as well as average 1-10 Likert scores from post-session evaluation. We see that 
TMR is strongly correlated with a round being not comprehensible ($\rho=0.78$).}
\label{tab:full-human-eval}
\end{table*}

\subsection{Data Collection}
\label{sec:data-collection}
\paragraph{Per-Turn Evaluation} After each turn, participants were shown a transcript of the model's previous response and asked to review it word by word (see Figure \ref{fig:annotation-interface}). They were instructed to highlight any spans of text that they did not understand, which we later used to compute TMR. Participants were also asked to report whether they felt that they could understand the overall meaning of the response. We used these annotations to compute overall statistics for comprehensibility.

\paragraph{Per-Conversation Evaluation}
In addition to annotating each response, participants completed a brief questionnaire after each full conversation, consisting of the following 10-point Likert scale questions:
\begin{enumerate}[noitemsep,topsep=5pt,leftmargin=*]
    \item How much of the bot could you understand?
    \item How effortful was it to talk to the bot?
    \item How comfortable did you feel talking to the bot?
    \item Did you feel like the bot's responses were natural given the context of the conversation?
    \item Would you want to chat with this version of the bot again in the future?
\end{enumerate}
The full text of the evaluation form can be found in Appendix \ref{sec:human-study-details} along with the full intake form, consent form, and list of conversation topics.

\subsection{Findings}
\label{sec:human-study-results}
\paragraph{FUDGE substantially improves comprehensibility over baseline (39.4\% -> 83.3\%)}
In Table \ref{tab:full-human-eval} we report the aggregated results of our human evaluation. FUDGE achieves roughly half the TMR of the other methods and was rated as the easiest to understand (7.67) and as taking the least amount of cognitive effort to speak with (4.67).\footnote{For full questionnaire results, see Table \ref{tab:human-eval-results} in the Appendix} We also see that FUDGE maintains a high degree of fluency, being rated the second most natural sounding (7.33) after the overgenerate method (7.50).

\paragraph{Prompting suffers from ``alignment drift''}
in Figure \ref{fig:4panel} we show plots of TMR and comprehensibility across subsequent rounds of conversation. Like \citet{almasi2025alignmentdriftcefrpromptedllms} we find that prompt-based methods for difficulty control suffer from \textbf{``alignment drift''}---gradually straying more and more from the target difficulty as the conversation progresses (see Table \ref{tab:full-conversation} for a full example conversation). We find that FUDGE exhibits consistently low TMR throughout multiple rounds and does not meaningfully drift upward over time.

\paragraph{TMR exhibits strong correlation with human judgements of quality}
Despite its inherent simplicity, we find that TMR exhibits strong correlation with human judgements for comprehensibility ($\rho=0.78$) as compared to other metrics like JReadability ($\rho=-0.17$) and ControlError ($\rho=0.40$). This result is promising for future attempts to improve performance on this task, as costly human evaluation may be avoidable in lieu of automatic computation of TMR.

\begin{figure}[t]
    \centering
    \includegraphics[width=\columnwidth]{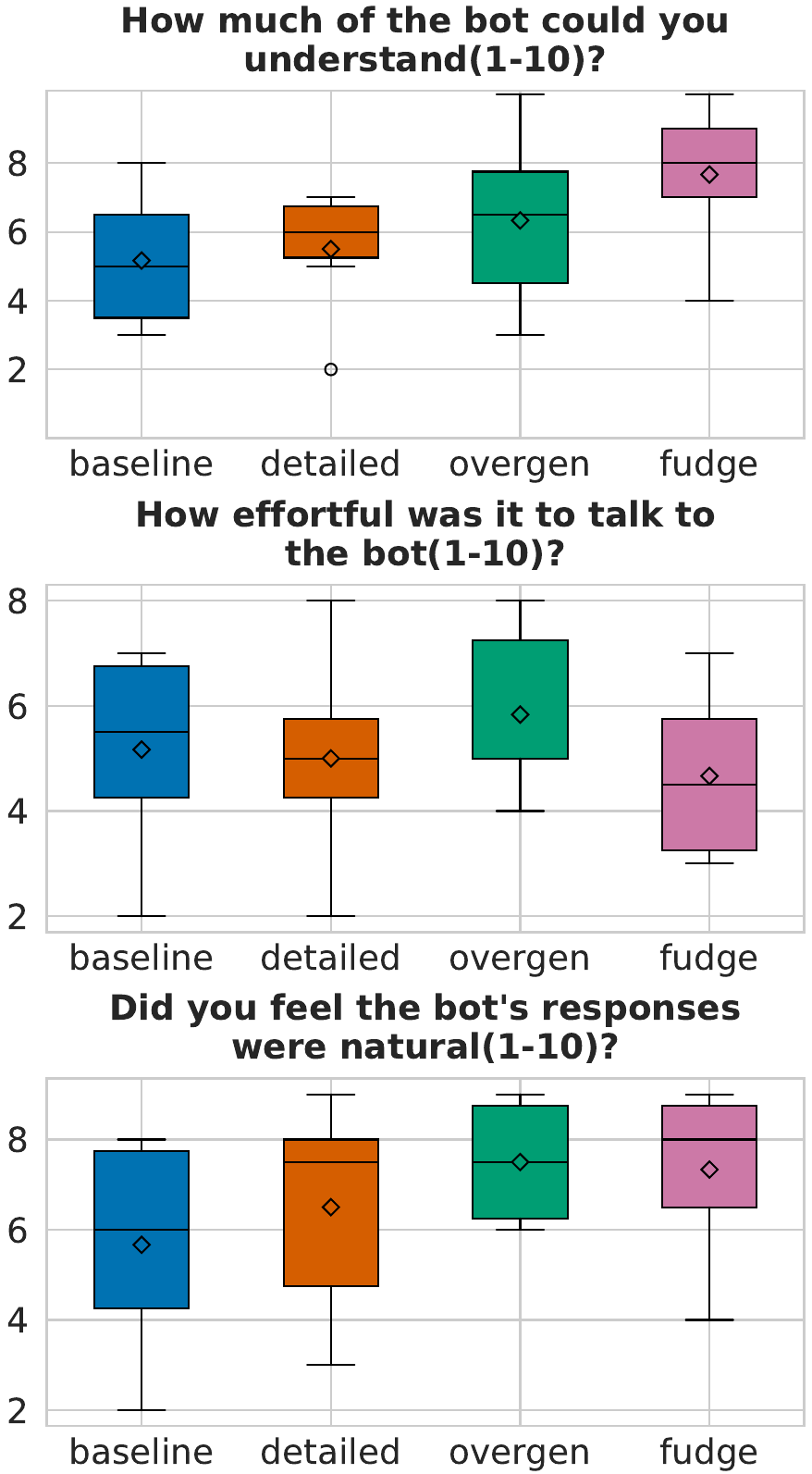}
    \caption{Box-and-whisker plots for selected questions from our post-conversation questionnaire (\S\ref{sec:data-collection}). We see that while FUDGE scores highest on average, there is substantial variance across methods.}
    \label{fig:three_boxplots}
\end{figure}

\paragraph{Human judgements exhibit high variance}
In Figure \ref{fig:three_boxplots} we report box-and-whisker plots for the distribution of questionnaire scores for each method. While in aggregate we see that FUDGE is the top performer, there is still substantial variance between different methods. For example, the detailed prompt showed the least variability in comprehension, with almost every participant rating it between 5 and 7. On the other hand, the Overgenerate method had much higher variance, with a low score of 3 and a high of 10. Given that a single incomprehensible turn can potentially derail a conversation, there is lots of room for improvement when it comes to the consistency of our difficulty control methods. Future work should focus on improving this in order to fully reap the benefits of fully fluent and natural AI conversation partners.

\section{Conclusion}
Learning a second language can open up significant opportunities. However, many beginner students find it difficult to practice conversational speech due to a lack of available conversation partners. In this paper we demonstrate the feasibility of using controllable generation techniques along with state-of-the-art large language models as AI conversation partners for learning Japanese.

Through both human and automatic evaluations we show that using controllable generation techniques from \citet{yang-klein-2021-fudge} allows us to substantially constrain the difficulty of chatbot responses with over 83\% of utterances rated as understandable by beginner speakers (JLPT N4-N5; CEFR A1-A2). We also show that this result stays consistent across multiple rounds and that the outputs of the bot are natural.

Unlike prior work, the best-performing technique in our evaluation does not require any fine-tuning of the base model, allowing for more personalized difficulty control. We imagine a scenario where students each have their own personal on-device predictor models trained to predict exactly their proficiency level. During conversation practice, these on-device predictors can combine with logits sent from an API-based chatbot for client-side difficulty control and personalization. We believe that our results constitute a meaningful step towards democratizing language learning and making conversation practice more accessible to speakers of all levels.

\section*{Limitations}

\paragraph{Single Language Focus} We acknowledge that conducting the study only for Japanese learners limits generalizability. We chose Japanese primarily because it allowed the authors to reliably sanity-check and review outputs during the complex development of a novel metric (TMR) and human study interface. We recognize that language-specific features—such as those related to morphology, tokenization, or syntax—may affect how modular control methods perform. We see expanding to other languages as an important next step, and it is our hope that our work can inspire researchers with resources and linguistic knowledge in other languages to explore this task more broadly.

\paragraph{Definition of Difficulty}
It is hard to define an one-off approach to measure the level of difficulty of an utterance. Difficulty can be based on vocabulary level, sentence length, grammatical complexity, or readability scores, etc. Our criteria chosen or way of implementation may not perfectly align with a target learner's actual proficiency or capture the full nuance of language difficulty. However, despite this inherent subjectivity and potential misalignment, defining measurable difficulty criteria is a necessary first step to enable any form of automated difficulty control or evaluation, even if the criteria themselves are imperfect representations of human difficulty perception.

\paragraph{Undetected Tokens in TMR}
The Token Miss Rate is computed as the ratio of tokens detected in ``above'' levels to the total count of all tokens in the utterance. This calculation implicitly classifies tokens that were unable to be binned to a difficulty level as being understood by the learner. This can potentially skew the score for utterances with a high proportion of undetected tokens.

\paragraph{TMR Relies on Exact Token Matching} We acknowledge that the TMR relies on exact token matching against predefined vocabulary lists, which is a simplification of true language comprehension. While we found that TMR correlates strongly with human comprehension judgments ($\rho = 0.78$, Table \ref{tab:human-eval-results}), it does not capture all nuances of understanding. In particular, not every missed token has an equal impact: two responses with the same TMR might differ significantly in overall comprehensibility, depending on which specific tokens are missed and how central they are to conveying meaning. Future work should explore more sophisticated evaluation methods, such as semantic-weighted variants of TMR or measures that incorporate context-aware comprehension to better capture true comprehension.

\section*{Acknowledgments}
We would like to thank the members of the University of Pennsylvania Japanese language department (JPAN), in particular Tomoko Takami and Nana Takeda-Kolb, for their enthusiasm and cooperation with our study. We would also like to thank the students of JPAN0100 and JPAN0210 for their participation and thoughtful feedback. In particular Andrew Zhu and Annika Heuser were particularly important for their early and consistent iterative feedback on the interface and qualitative evaluations of model outputs.

We would also like to thank the members of the PennNLP group, in particular Alyssa Hwang, Neil Sehgal, Allen Chang, Josh Ludan, Alon Jacoby, and Hita Kambhamettu for their feedback on early versions of the draft---without whom the clarity of writing would be substantially worse.

This research is supported in part by the Office of the Director of National Intelligence (ODNI), Intelligence Advanced Research Projects Activity (IARPA), via the HIATUS Program contract \#2022-22072200005. The views and conclusions contained herein are those of the authors and should not be interpreted as necessarily representing the official policies, either expressed or implied, of ODNI, IARPA, or the U.S. Government. The U.S. Government is authorized to reproduce and distribute reprints for governmental purposes notwithstanding any copyright annotation therein.

\bibliography{custom}

\newpage
\appendix

\section{Full Details of the FUDGE Method}
\label{sec:fudge}

In this section we give a more detailed explanation of the FUDGE method \citet{yang-klein-2021-fudge} and provide more detail about our training setup for the predictor model.

\subsection{Controllable Text Generation}
Controllable text generation is the task of generating text $X = {x_1\ldots x_n}$ conditioned on a specific desired attribute $a$ (e.g. the desired difficulty of an utterance). Let $\mathcal{G}$ be an LLM which models $P(x_i|x_{1:i-1})$ for tokens $x_1\ldots x_n$. The likelihood of the full sequence $P(X)$ can be factorized as
\[P(X) = \prod_{i=1}^{n} P(x_i|x_{1:i-1})\]
To condition on a particular attribute, we must instead model $P(X|a)$, which modifies the previous factorization:
\[P(X|a) = \prod_{i=1}^{n} P(x_i|x_{1:i-1}, a)\]
Thus the controlled generation task can be simplified to finding a method for modeling the attribute-conditional distribution $P(x_i|x_{1:i-1}, a)$.

\subsection{Future Discriminators (FUDGE)}
FUDGE \cite{yang-klein-2021-fudge} operates by approximating $P(x_i|x_{1:i-1}, a)$ based on a Bayesian factorization:
\[P(x_i|x_{1:i-1}, a) \propto P(a|x_{1:i}) \cdot P(x_i|x_{1:i-1})\]
Where $P(a|x_{1:i})$ denotes the likelihood of the attribute $a$ given the prefix $x_{1:i}$. Intuitively this can be seen as a re-ranking of the candidate tokens from $\mathcal{G}$ by some attribute predictor model $\mathcal{M}$.

Importantly, this technique requires the predictor model $\mathcal{M}$ to model how likely it is that a particular sequence will have the attribute \textit{in the future}. This is a very different task from the standard sequence classification task, thus preventing us from using off-the-shelf classifiers as predictors.

\subsection{Training Predictors for Difficulty}
\label{sec:training-predictors}
Given a dataset $D$ of text-attribute tuples $(x_{1:n}, a')$ the predictor model $\mathcal{M}$ is fine-tuned using all prefix-attribute pairs $(x_{1:i}, a')$ to learn the \textit{future} likelihood of the attribute given the prefix. We modify the predictor to minimize a multi-class cross entropy loss instead of a binary loss.
\[\mathcal{L}_{\mathcal{M}} = -\sum_{x}\log P(a'|x)\]
This encourages the model to output a distribution of likelihoods over $n$ difficulty levels.

\begin{table}[t]
\centering
\setlength{\tabcolsep}{5pt}
\renewcommand{\arraystretch}{1.05}
\begin{tabular}{@{}l c@{}}
\toprule
\textbf{Hyper‑parameter} & \textbf{Value} \\ \midrule
per\_device\_train\_batch\_size & 16 \\
per\_device\_eval\_batch\_size  & 16 \\
num\_train\_epochs             & 3 \\
learning\_rate                 & $5\times10^{-5}$ \\
evaluation\_strategy           & epoch \\
logging\_steps                 & 10 \\
dataloader\_num\_workers       &  4 \\
\bottomrule
\end{tabular}
\caption{Hyper‑parameters used to train the FUDGE predictor model. }
\label{tab:bp-training-config}
\end{table}

\subsection{Inference}
\label{sec:inference}
To re-rank candidate tokens for a particular prefix $x_{1:i-1}$ the predictor model is run on each potential next token sequence $x_{1:i}$. Since doing this for the full distribution would be prohibitively expensive, we follow \citet{yang-klein-2021-fudge} and truncate the next token distribution to top-$k$=50 sampling. To obtain the logit vector for the predictor we select the logit at the index corresponding to the target difficulty for each of the top-$k$ tokens.

We also employ a control parameter $\lambda$ which allows us to modify the magnitude of the difficulty predictor. The logit vector $\hat{\textbf{y}}$ for the full system can thereby be calculated by computing:
\[\hat{\textbf{y}} = \lambda \textbf{a} + (1 - \lambda) \textbf{x}\]
where $\textbf{x}$ and $\textbf{a}$ are the truncated logit vectors from the LLM and the predictor respectively. After this we normalize the vector $\hat{\textbf{y}}$ and sample the next token as usual.

\subsection{Training the Predictor Model}
\label{sec:hyperparams}
We fine-tune the \texttt{answerdotai/ModernBERT-base} transformer on sentence-level JLPT annotations from the \textbf{jpWaC-L} corpus \cite{erjavec-etal-2008-jpwac} to predict the level of a sentence given a partially-complete prefix. That is, for a given sentence $x_{[1:n]}$, the predictor is trained on all prefixes $\{x_{[1]}, x_{[1:2]}, \cdots, x_{[1:n-1]}, x_{[1:n]}\}$ to predict the difficulty level of the full sentence $x_{[1:n]}$. 

To address the class imbalance present in our corpus (see Table~\ref{tab:jpwac-distribution}), we downsample each JLPT level to the size of the smallest level, ensuring equal representation across levels (N1–N5). This produces a balanced dataset without over‑ or under‑representing any proficiency level. All relevant hyperparameters and settings are summarized in Table~\ref{tab:bp-training-config}. We trained our model using one NVIDIA A6000 GPU for 15.5 hours and our trained model can be found and downloaded from Huggingface.\footnote{\url{https://huggingface.co/emmajin0210/modernbert_output_endtok_readable}}

\begin{table}
\begin{CJK}{UTF8}{goth}
\centering 
\small
\begin{tabular}{p{0.3\linewidth}|p{0.2\linewidth}} 
\toprule
\textbf{JLPT N5} & 36563 \\ 
\midrule
\textbf{JLPT N4} & 103298 \\
\midrule
\textbf{JLPT N3} & 372421 \\
\midrule
\textbf{JLPT N2} & 137312 \\
\midrule
\textbf{JLPT N1} & 2627335\\
\midrule
\textbf{Total} & 3276929 \\
\bottomrule
\end{tabular}
\caption{Total number of sentences per JLPT level in the jpWaC corpus \cite{erjavec-etal-2008-jpwac}. }
\label{tab:jpwac-distribution}
\end{CJK}
\end{table}

\section{Computing TMR from Heuristic Vocabulary Bins}
\label{sec:appendix_overgen}
When computing TMR for the reranking process in the overgenerate method, we construct our vocabulary bins using heuristics derived from the \textbf{jpWaC-L} corpus \cite{erjavec-etal-2008-jpwac} as follows:

\paragraph{Step 1: Tokenization.}
For each collection of sentences belonging to a certain JLPT level in the \textbf{jpWaC-L} corpus \cite{erjavec-etal-2008-jpwac}, we tokenize every line by using Sudachi \cite{takaoka-etal-2018-sudachi} in coarse-grained mode~C. Tokenization includes lemmatization and punctuation removal, and we additionally filter for strings consisting entirely of Japanese script (kanji, hiragana, katakana) using a Unicode whitelist.

\paragraph{Step 2: Rare Word Filtering.}
To eliminate spurious and rare tokens, we apply two threshold filters:
\begin{itemize}[noitemsep, topsep=0.3em,leftmargin=1.3em]
    \item \textbf{Global frequency filter:} token must consist of more than $10^{-6}$ of all tokens in all levels
    \item \textbf{Level-specific filter:} token must be more than $10^{-6}$ of the tokens within their level.
\end{itemize}
These thresholds eliminate rare singleton tokens and help smooth over any typos or other one-off issues in the original data.

\paragraph{Step 3: Best-Level Assignment.}
We assign each token $w$ to the easiest level in which it reaches a sufficient relative frequency. That is, for each token, we scan levels N5→N1 in order and assign it to the first level where its frequency is non-trivial relative to that level’s total token count:
\[
  \text{score}_{w, \ell} = \frac{\text{count}_{w,\ell}}{\text{total tokens}_\ell}
\]
The first level $\ell$ for which $\text{score}_{w,\ell}$ exceeds a small threshold is selected as the word’s bin.

\section{Computing TMR from Anki Decks}
\label{sec:vocabulary-bins}
When computing TMR for evaluation, rather than construct vocabulary lists using imperfect heuristics, we opt to construct a set of gold standard JLPT vocabulary bins from official study material. To do this, we directly parse flashcards from the \texttt{jlpt-anki-decks} dataset \cite{chyyran_jlpt_anki}.

\paragraph{Input Format.}  
Each flashcard deck is provided in the form of an `.apkg' file. These decks are organized by JLPT level (N5–N1) and contain a set of Japanese expressions, readings, and glosses for each level. We import each deck into Anki\footnote{\url{https://apps.ankiweb.net/}} and store it as a local SQLite file under `collection.anki2'. Each vocabulary entry is stored as a row in the `notes' table, where the `flds' field contains tab-separated content fields, including: Expression (kanji, kana, or mixed), Reading (optional), Gloss/Meaning (in English). 

\begin{table*}[t]
\begin{CJK}{UTF8}{goth}
\centering 
\small
\begin{tabular}{p{0.08\linewidth}|p{0.86\linewidth}} 
\toprule
\textbf{JLPT N5} & 明日 (tomorrow), あなた (you), 魚 (fish), いいえ (no, not at all), 少し (little, few), 有名 (famous) \\ 
& 先生 (teacher, professor; master; doctor), 有る (to be, to have), 会う (to meet, to see),  行く (to go) \\
\midrule
\textbf{JLPT N4} & 生きる (to live), 心配 (worry, concern), 始める (to start, to begin), 止める (to end, to stop) \\
& 中学校 (junior high school), 会話 (conversation), そろそろ (gradually, soon), 専門 (major; speciality) \\
\midrule
\textbf{JLPT N3} & 様々 (varied, various), 全て (all, the whole, entirely), 集まり (gathering, meeting, collection) \\
& 印象 (impression), 作品 (work, opus, production), わざと (on purpose), 変化 (change, variation, shift) \\
\midrule
\textbf{JLPT N2} & 重力 (gravity), 純粋 (pure, genuine, unmixed), 先祖 (ancestor), 課税 (taxation), 清い (clear, pure, noble) \\
& 強化 (strengthen, intensify, reinforce), 論ずる (to argue, to discuss), 必需品 (necessities, essential) \\
\midrule
\textbf{JLPT N1} & 賢明 (wisdom, intelligence, prudence), 倹約 (thrift, economy, frugality), 鉱業 (mining industry) \\ 
& 護衛 (guard, convoy, escort), 戸籍 (census, family register), 臆病 (cowardice, timidity), 放射能 (radioactivity) \\
\bottomrule
\end{tabular}
\caption{Example vocabulary at each JLPT level as parsed from the \texttt{jlpt-anki-decks} dataset \cite{chyyran_jlpt_anki}.}
\label{tab:jlpt-bins}
\end{CJK}
\end{table*}

\paragraph{Normalization Heuristics.}  
\begin{CJK}{UTF8}{goth}
For each entry we apply the following normalization steps.
\begin{itemize}[itemsep=0.1pt,topsep=0.5em]
    \item Japanese parentheses are first converted to standard parentheses ( and ).
    \item Tilde-like characters (e.g., 〜) are stripped entirely, as they denote alternation or ellipsis.
    \item Parenthetical expressions (e.g. (を), (する)) are expanded to both the outside form (e.g., 話す (こと) becomes 話す) and the full form (e.g., 話すこと)
    \item Readings are similarly expanded, and matched against expressions using filters to avoid over-generation.
\end{itemize}
\end{CJK}

\paragraph{Filtering Strategy.}  
We also include regular expression-based heuristics to exclude non-informative entries:
\begin{CJK}{UTF8}{goth}
\begin{itemize}[itemsep=0.2pt,topsep=0.5em]
    \item Skip readings that are only one hiragana character (e.g., の), or extremely short forms unlikely to be useful alone.
    \item Avoid adding a reading as a standalone entry if it overlaps directly with another alternative expression.
    \item Skip duplicates, malformed entries, or entries missing either expression or meaning.
\end{itemize}
\end{CJK}
\paragraph{Output Format.}  
The result is a set of per-level JSON files, e.g., \texttt{n5.json}, each containing normalized entries in the form:
\begin{lstlisting}[basicstyle=\ttfamily\small, escapeinside={(*@}{@*)}]
{
  (*@\begin{CJK}{UTF8}{goth}"会う"\end{CJK}@*): {
    "meaning": "to meet, to see"
  },
  (*@\begin{CJK}{UTF8}{goth}"あなた"\end{CJK}@*): {
    "meaning": "you"
  },
  ...
}
\end{lstlisting}
In Table~\ref{tab:jlpt-bins}, we show a representative sample of vocabulary items used at each JLPT level from N5 (easiest) to N1 (hardest), along with their English glosses. We provide the full list of vocabulary in our code repository.

\section{Web Interface \& Implementation}
\label{sec:web-interface}
\begin{figure*}
    \centering
    \begin{subfigure}[t]{\columnwidth}
        \centering
        \includegraphics[width=\columnwidth]{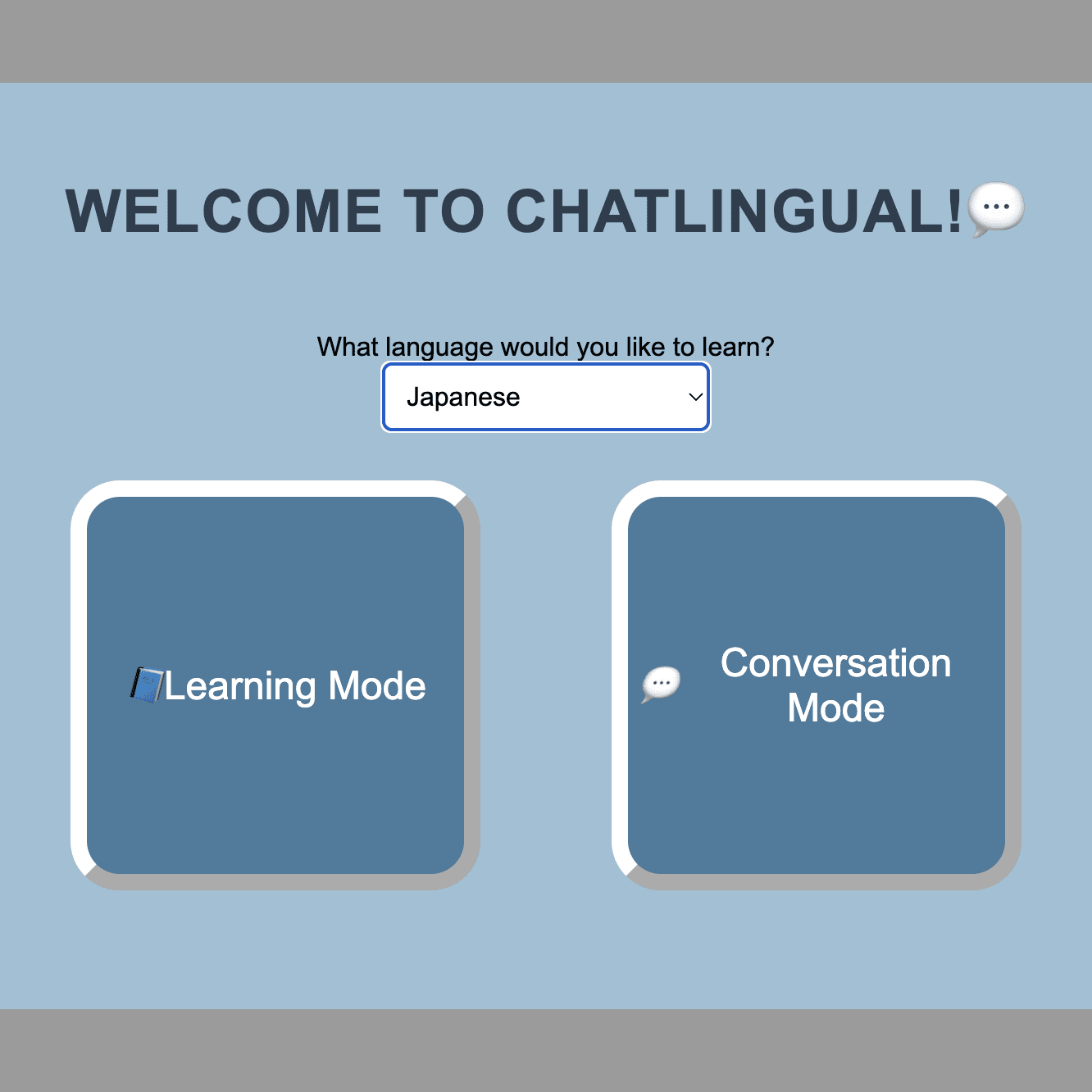}
        \caption{The \textbf{homepage} of the interface used for the human evaluation. For the purpose of this study, we only use the Conversation Mode.}
        \label{fig:interface-homepage}
    \end{subfigure}
    ~
    \begin{subfigure}[t]{\columnwidth}
        \centering
        \includegraphics[width=\columnwidth]{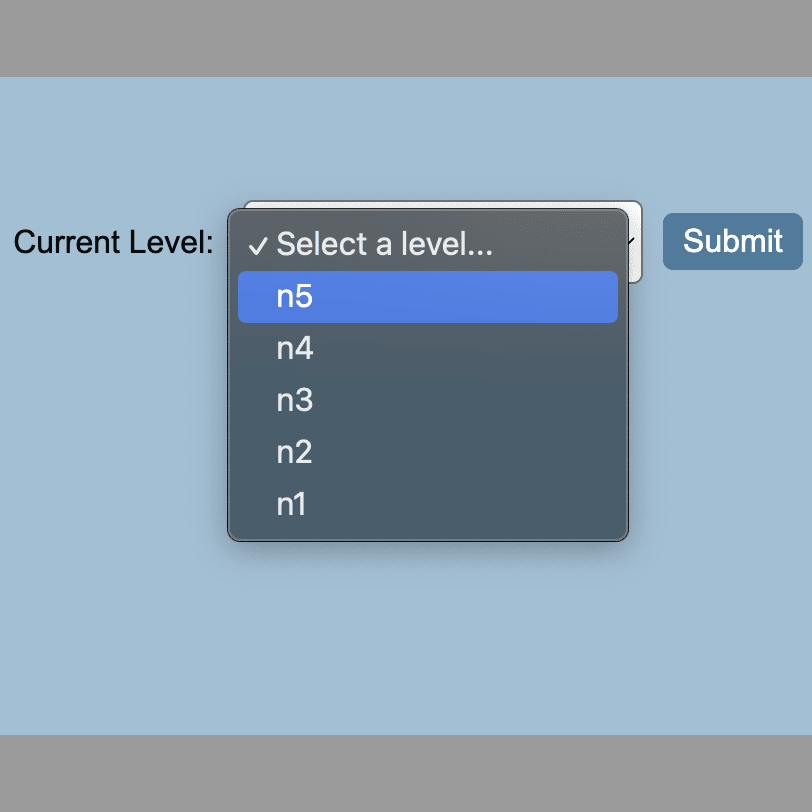}
        \caption{We select the \textbf{user's corresponding level} through a dropdown.}
        \label{fig:interface-select-level}
    \end{subfigure}
    
    \begin{subfigure}[t]{\columnwidth}
        \centering
        \includegraphics[width=\columnwidth]{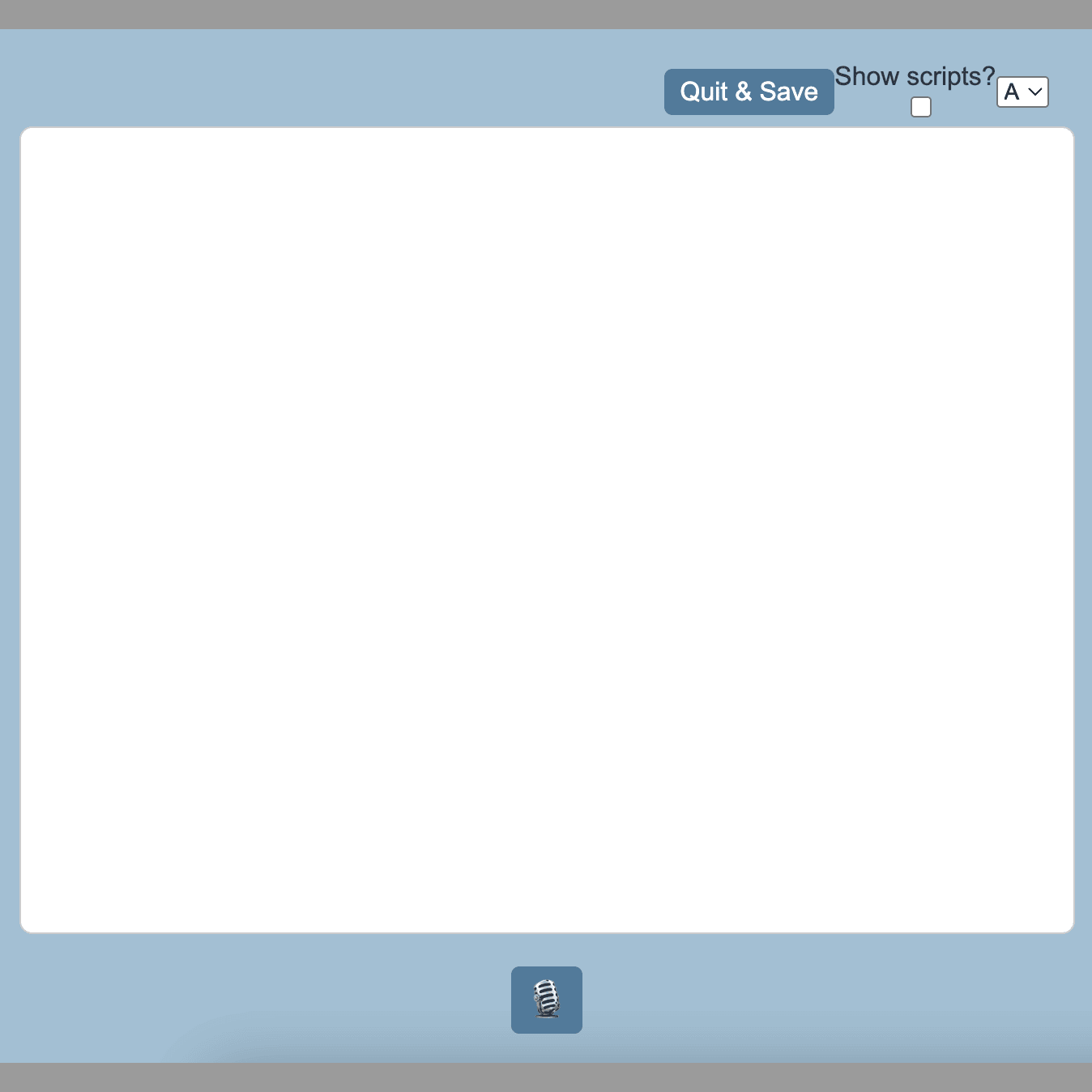}
        \caption{The user interacts with the \textbf{main chat interface}, which is initially blank with a recording button.}
        \label{fig:interface-chat}
    \end{subfigure}
    ~
    \begin{subfigure}[t]{\columnwidth}
        \centering
        \includegraphics[width=\columnwidth]{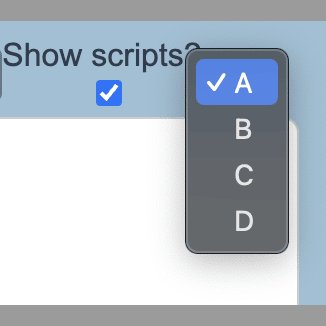}
        \caption{We select the \textbf{method} through a dropdown before each round; we also set \textbf{show scripts} to true so that the user can read the text output of the bot. }
        \label{fig:interface-select-method}
    \end{subfigure}

    \caption{Screenshots from our web interface along with the procedure used during each user study round. Interface views are: Homepage (a), Level-Selection (b), Chat Interface (c), and Method-Selection (d).}
    \label{fig:interface-prep}
\end{figure*}

For our human evaluation we build a custom web interface for participants to interact with. Figure~\ref{fig:interface-prep} presents the stages of preparation for each round of user study, including navigating from the homepage (a), selecting the corresponding JLPT level of the user (b), navigating to the chat interface (c), and selecting which method to run on (d). Participants were able to speak aloud their responses which were transcribed with the API-based \texttt{whisper-1} model \cite{radford2022whisper} and given as input to the backend model via a FastAPI query. The model's response was read aloud with audio generated by the OpenAI TTS-1 model.\footnote{\url{http://platform.openai.com/docs/models/tts-1}}

\paragraph{Kani}
For our chatbot application logic we use the Kani\footnote{\url{https://github.com/zhudotexe/kani}} framework, which is a flexible abstraction for building multi-turn conversational agents in Python \cite{zhu-etal-2023-kani}. It provides lightweight primitives around LLM engines (\texttt{BaseEngine}), message formats (\texttt{ChatMessage}, \texttt{ChatRole}), and generation management (\texttt{Completion}). This allows us to easily swap between local and API based chat models without changing our interface code. We implemented FUDGE using the HuggingFace \texttt{LogitsProcessor} interface which allowed us to apply our predictor to any Huggingface model that shares the Qwen tokenizer.

\paragraph{Generation Parameters}
For all models we used the default parameter settings as listed in Huggingface and the OpenAI API. For Qwen these parameters are temperature=0.7, top\_p=0.8, top\_k=20, and repetition\_penalty=1.05. For GPT-4 this (likely)\footnote{Unlike the Completions endpoint, the ChatCompletions endpoint in the OpenAI API does not actually specify the default values for temperature and top\_p nor does it return them on successful completion. Thus we list the defaults for the Completions API here in the hopes that they are the same.} is temperature=1.0, top\_p=1.0. For the student model in the ``self-chat'' pipeline we explicitly set temperature=0.7 and top\_p=1.0 for both Qwen and GPT-4 to promote higher diversity of response. Neither of the models were queried with ``reasoning'' turned on. Future work should investigate how the addition of a thinking token budget may change our results, particularly around the alignment drift phenomenon.

\paragraph{Hardware}
For inference we used six NVIDIA RTX A6000 GPUs queried via local server through a web socket. For FUDGE, we allocated one GPU to the predictor model and the other five GPUs to host the 72 billion parameter Qwen model. For cases during ``self-chat'' where a student Qwen agent was chatting with a tutor Qwen agent, we loaded one instance of the model and swapped out the chat context as necessary.

\section{Human Study Details}
\label{sec:human-study-details}

\subsection{Intake Form}

After signing up and prior to their user study session, each participant completed an intake form. This form collected language background, class enrollment, and JLPT experience, along with confirming informed consent for participation and recording. The full content of the intake and consent form is shown in Figure~\ref{fig:intake-form-appendix}.

\label{sec:intake_form}
\begin{figure*}[ht]
  \centering
  \begin{lstlisting}[basicstyle=\ttfamily\small, escapeinside={(*@}{@*)}]
(*@\textbf{Intro:}@*)
Thank you again for signing up to participate in our study!

This is a small-scale user study, so your participation truly means a lot to us. 
The information you provide here will help us schedule your session, accommodate your preferences, and better understand your language background.

All responses will be kept confidential and used solely for research purposes.
We are collecting your email only to coordinate your session and follow up if needed. It will not be shared or used for any other purpose.


(*@\textbf{Informed Consent:}@*)

1. Informed Consent Statement:
This study involves interacting with an AI-powered chatbot in Japanese. While we've designed it to be level-appropriate, 
the chatbot may occasionally provide inaccurate or confusing responses. 
Your interactions will be recorded and anonymized for research purposes. 
Participation is voluntary, and you may withdraw at any time without penalty.

All data collected will be kept confidential and used solely for research purposes. 
No personally identifiable information will be shared outside the research team. 
Audio from your session will be used to analyze the interaction and will be deleted after it is transcribed and anonymized.

Consent Confirmation:
By continuing, you confirm that:
- You are at least 18 years old
- You have read and understood the information above
- You voluntarily consent to participate in this study

[ ] I consent to participate in this study under the terms described above.


(*@\textbf{Participant Background:}@*)

1. Did you grow up speaking or hearing Japanese at home as a child?
[ Yes / No ]

2. Approximately how many hours per week do you spend speaking Japanese conversationally outside of class?
[ Text response ]

3. Approximately how many hours per week do you spend listening to conversational Japanese (podcasts, etc.) outside of class?
[ Text response ]

4. If you're currently taking a Japanese class at the university, which class are you enrolled in?
[ Text response ]

5. How many semesters of Japanese study have you completed?
[ Text response ]

6. Have you taken the Japanese Language Proficiency Test (JLPT)?
[ Yes / No ]

If yes, what level did you achieve?
[ N5 / N4 / N3 / N2 / N1 ]

If not, what level do you estimate you're at (based on textbook, course, or self-assessment)?
[ N5 / N4 / N3 / N2 / N1 ]
  \end{lstlisting}
  \caption{Full content of the form used to collect intake and consent data in the study.}
  \label{fig:intake-form-appendix}
\end{figure*}

\begin{table*}
\begin{CJK}{UTF8}{goth}
\centering 
\small
\begin{tabular}{p{0.7\linewidth}|p{0.1\linewidth}} 
\toprule
\textbf{Genki I}, \textbf{Genki II} (Lesson 13-Lesson 14) & JLPT N5 \\ 
\midrule
\textbf{Genki II}, \textbf{Tobira: Gateway to Advanced Japanese} (Unit 1–Unit 2) & JLPT N4 \\
\midrule
\textbf{Tobira: Gateway to Advanced Japanese} (Unit 9-Unit 14) & JLPT N3 \\
\midrule
\textbf{Advanced materials selected from the internet, newspapers, and books} & JLPT N2 \\
\bottomrule
\end{tabular}
\caption{Mapping from textbooks used in university Japanese courses to their corresponding JLPT levels. We used this mapping to assign students to JLPT levels based on their (self-reported) university course progress.}
\label{tab:textbook-to-jlpt}
\end{CJK}
\end{table*}

\begin{table*}[t]
\centering
\setlength{\tabcolsep}{3pt}
\small
\begin{tabular}{lccccc}
\toprule
\multicolumn{6}{c}{\textbf{Human Evaluation -- Exit Survey Full Results}}\\
\midrule
\textbf{Model} &\textbf{Comfort?} $\uparrow$&\textbf{Understandable?} $\uparrow$& \textbf{Effortful?} $\downarrow$& \textbf{Natural?} $\uparrow$& \textbf{Chat Again?} $\uparrow$\\
\midrule
Baseline (Qwen)& 7.33 & 5.17 & 5.17 & 5.67 & 7.33\\
Detailed (Qwen)& \textbf{8.00} & 5.50 & 5.00 & 6.50 & 7.33\\
Overgenerate (Qwen)& 6.83 & 6.33 & 5.83 & \textbf{7.50} & 7.33\\
FUDGE ($\lambda = 0.8$) (Qwen)& 6.83 & \textbf{7.67} & \textbf{4.67} & 7.33 & \textbf{7.67}\\
\bottomrule
\end{tabular}
\caption{Full results of the exit survey given to our participants. Each question was graded on a 10-point Likert scale. We see that the detailed prompt and overgenerate methods scored highest on comfort and naturalness while FUDGE scored best on difficulty control metrics.}
\label{tab:human-eval-results}
\end{table*}

\subsection{Mapping of Textbook to JLPT level}
\label{sec:textbook_to_jlpt_mapping}

To estimate a participant’s JLPT level, we mapped their current university Japanese course (as self-reported) to an approximate JLPT level based on textbook progression and kanji coverage. Table~\ref{tab:textbook-to-jlpt} summarizes this mapping, which uses both Genki and Tobira textbooks as primary anchors for N5 through N3 levels, and more advanced materials such as news articles to the N2 level.

\begin{table*}
\centering 
\small
\begin{tabular}{p{0.08\linewidth}|p{0.86\linewidth}} 
\toprule
&introduce yourself (such as your name, job/school, where you're from, etc.)\\
\textbf{JLPT N5}&describe what you usually do in the morning and evening\\
&talk about your favorite food and where you usually eat it\\
\midrule
&explain what you will do this weekend and with whom\\
\textbf{JLPT N4}&describe your favorite hobby and how often you do it\\
&talk about a typical day at school or work, including schedule and people you meet\\
\midrule
&describe a travel experience: where you went, what you saw, and who you went with\\
\textbf{JLPT N3}&talk about planning a birthday party: location, food, and guests\\
&describe your favorite movie: the story, characters, and why you like it\\
\midrule
&describe a recent news story you found interesting, and why it caught your attention\\
\textbf{JLPT N2}&explain one cultural difference between Japan and your country, and how it affects communication\\
&discuss a challenge people face when communicating in a Japanese workplace\\
\midrule
&discuss recent advancements in regenerative medicine and their ethical implications in Japan\\
\textbf{JLPT N1}&explain the role of quantum computing in future communication technologies and how Japan is preparing for it\\
&analyze the impact of declining biodiversity on Japan's agricultural sustainability and food security\\
\bottomrule
\end{tabular}
\caption{The full list of conversation topics used in our self-chat evaluation pipeline (see Section \ref{sec:evaluation-pipeline}). During evaluation, each method generated one conversation for each of the fifteen topics listed. This was done to remove any semantic bias that could stem from randomly sampling topics from a larger pool.}
\label{tab:topics}
\end{table*}

\subsection{Conversation Topics}
\label{sec:human_study_topics}
In Figure \ref{fig:jlpt-topic-list} we list the set of level-appropriate conversation topics that we compiled based on common study materials and university textbooks. Before each round, our system randomly selects three topics from the corresponding list and presents them for the participant to choose from. We allow our participants to choose from among the three topics for each new conversation they undergo and do not allow duplicate topics within the same session. We do this to ensure some minimum level of comfort in speaking about a particular topic.

\subsection{Full Survey Results}
The full results of our exit survey are reported in Table \ref{tab:human-eval-results}. We see that FUDGE scores the highest on difficulty control metrics such as understandability and effort while the detailed prompt and overgenerate methods scored highest on fluency metrics such as comfort and naturalness.

\begin{figure*}[t]
\centering
\begin{lstlisting}[basicstyle=\ttfamily\small,  escapeinside={(*@}{@*)}]
(*@\textbf{JLPT N5 Topics List:}@*)
- Summer vacation plans
- Weekend hobbies or routines
- Favorite movie or TV show
- Favorite book, story, or folklore
- Favorite sport or physical activity
- A memorable trip or vacation
- A time you got sick
- A favorite holiday or festival

(*@\textbf{JLPT N4 / Early N3 Topics List:}@*)
- A time something was stolen
- A time you were hurt or injured
- Doing house chores
- A habit that annoys you
- A time you reported a crime or accident
- A favor you asked from someone
- A time you had to say goodbye
- A promise or decision you made
- A future goal that you have
- A region in Japan you want to visit
- A famous place you've been to
- A local food or specialty you like
- A festival you've attended or want to see
- Your hometown and what it's known for
- A memorable travel story
- A seasonal event you enjoy
- A travel recommendation for a friend

(*@\textbf{JLPT N2 Topics List:}@*)
- Describe a recent news story you found interesting, and why it caught your attention
- Explain one cultural difference between Japan and your country, and how it affects communication
- Discuss a challenge people face when communicating in a Japanese workplace
- Talk about a social issue you care about and why it's important to you
- Describe a time you had to be polite in a difficult situation
- Compare education systems in Japan and your home country
- Share your opinion on using AI or technology in daily life
- Describe a tradition or custom from your country and how it's changing
- Talk about how your communication style changes depending on the situation
- Discuss the pros and cons of working remotely or studying online
- Talk about a piece of Japanese literature you like
- Discuss how Japanese society is addressing the social issue of aging population
\end{lstlisting}
\caption{The pool of conversation topics used in the human evaluation. For each conversation, participants were able to choose a topic they felt comfortable with from the list associated with their estimated JLPT level. This was especially helpful to early beginner students who may only be able to speak on a very limited range of topics.}
\label{fig:jlpt-topic-list}
\end{figure*}

\section{System Prompts}
In Figure \ref{fig:tutor-system-prompt} we report the full text of the Baseline prompt and in Figure \ref{fig:detailed-tutor-system-prompt} we report the full text of the Detailed prompt. Both prompts are populated using information from Table \ref{tab:level-descriptions} with Baseline only getting the level word and Detailed getting all information. All JLPT level descriptions were sourced from the official JLPT level summaries webpage.\footnote{\url{https://www.jlpt.jp/e/about/levelsummary.html}} 

Additionally, in Figure \ref{fig:student-system-prompt} we report the full text of the student model's prompt. This prompt was held constant across all tutor methods tested and no additional difficulty control was applied to the student model.

\section{Example Conversations}
In Table \ref{tab:example-responses} we report an example dialogue turn between our human participants and each of the four methods tested. We also provide the participant's token-level comprehensibility annotations to help convey the extent to which each method is effectively controlling the difficulty of their response. While not all queries listed in Table \ref{tab:example-responses} were from the same student, they were all from students at the lowest level of proficiency (JLPT N5). From the table we can see that, even with our aggressive prompting and explicit difficulty control, language models still output a substantial number of vocabulary tokens that are above the target range.

In Table \ref{tab:full-conversation} we report the first four turns of a conversation between a single participant and the detailed prompt method. As we can see, the prompted model suffers from alignment drift and gradually begins ignoring instructions and outputting in its default style as the conversation progresses---especially once the student begins asking for recommendations. We suspect that post-trained models may have certain well-established response patterns (such as listing off recommended items) that, when triggered, accelerate the drift back to near-native difficulty.

Finally in Table \ref{tab:full-conversation-ai} we provide the first seven turns of a conversation from the ``self-chat'' pipeline. Despite the Tutor and Student model both being prompted and controlled to target the beginner level, we still get many tokens above JLPT N5. In addition, we note that the subjective quality and diversity of the conversation is quite different from the human setting---remaining rather formulaic and sterile throughout. This further reinforces the value of including rigorous human evaluations when studying difficulty control to properly take models away from standard response patterns.

\section{Use of AI Assistants}
AI assistants were used in writing this paper to enhance clarity of wording, format tables and figures, and help fix compilation errors. The models used were ChatGPT and GPT-o3.

\begin{table*}[t]
\begin{CJK}{UTF8}{goth}
\centering
\small
\begin{tabular}{lp{0.7\columnwidth}p{1.0\columnwidth}}
\toprule
&\textbf{Student Query}&\textbf{Model Response}\\
\midrule
Baseline&テレビと映画のほうがどちらが好きですか?\newline \textit{Which do you like more, TV or movies?}&私はテレビドラマが好きです。ストーリーが\textcolor{red}{ゆっくり進んで、キャラクターを深く知れるから}。あなたはテレビと映画、どちらが好きですか？\newline \textit{I like TV Dramas. The story \textcolor{red}{progresses more slowly so we can understand the characters more deeply.} Which do you like more, TV or movies?}\\
\midrule
Detailed&バキザグラブラは少年の名前はバキです。 とても強い人ですよ。 このストーリーはファイティングのトーナメントについてです。\newline \textit{In "Baki the Grappler" the boy's name is "Baki". He is very strong. The story is about a fighting tournament.}&そうですね、バキは強い少年で、ファイティングのトーナメントが\textcolor{red}{テーマなんですね。戦闘シーンが迫力満点だろうなと思います。}バキ以外にも好きなキャラクターはいますか？\newline \textit{Ahh I see, Baki is a strong boy, and a fighting tournament is the story's \textcolor{red}{theme. I'd imagine the battle scenes must be absolutely stunning.} Are the any other characters you like besides Baki?}\\
\midrule
Overgenerate&はい、まだまだまあまあえっと私は一週間ぐらい風邪をひったんでした。\newline \textit{Yes, I'm still feeling not great. I had a cold for about a week.}&そうですね、一週間くらい風邪を引いていたんですね。長い\textcolor{red}{期間}でしたね。ゆっくり\textcolor{red}{休んで回復することができてよかったです}。次に風邪を引いた時には、少し\textcolor{red}{早めに休む}のもいいかもしれませんね。\textcolor{red}{体調管理}、大切ですよ。\newline \textit{I see, you had a cold for about a week. That's quite a long \textcolor{red}{time period. I'm glad you were able to rest and recover}. Next time you catch a cold it might be good to start \textcolor{red}{resting a bit earlier}. \textcolor{red}{Health management} is very important after all.}\\
\midrule
FUDGE&こんにちは。夏休みに何をするつもりですか? \newline \textit{Hello. What are you planning to do for summer vacation?}&こんにちは！夏休みはまだ決めてないけど、友達と\textcolor{red}{海水浴}に行く\textcolor{red}{予定}です。あなたは夏休みに何をしたいですか？ \newline \textit{Hello! For summer vacation I still haven't decided yet but I do have \textcolor{red}{plans} to go \textcolor{red}{swim in the sea} with some friends. What would you like to do for summer vacation?}\\
\bottomrule
\end{tabular}
\caption{Example student queries and model responses from each of the four methods (Baseline, Detailed, Overgenerate, and FUDGE). Words and phrases highlighted in \textcolor{red}{red} were manually labeled as incomprehensible by a human participant. All queries represented in this figure were given by a participant who identified as ``absolute beginner'' level (JLPT: N5, CEFR: A1). We see that model responses are long and contain advanced phrases, even for simple queries, making conversation practice with chatbots cumbersome for beginner learners.}
\label{tab:example-responses}
\end{CJK}
\end{table*}

\begin{table*}
\begin{CJK}{UTF8}{goth}
\centering 
\small
\begin{tabular}{l|p{0.2\columnwidth}|p{0.55\columnwidth}|p{0.5\linewidth}} 
\toprule
&\textbf{Level Word}&\textbf{Guidelines}&\textbf{Description}\\
\midrule
\textbf{N5} & beginner &You should use only very basic vocabulary and simple sentence structures understandable in everyday situations.& - can understand only very basic Japanese\newline
- very easy expressions and sentences written in hiragana, katakana, and basic kanji\newline
- very short and easy conversations spoken slowly about topics regularly encountered in daily life and classroom situations\\ 
\midrule
\textbf{N4} & pre-intermediate &You should use simple grammar and vocabulary related to familiar daily topics, avoiding compound or abstract expressions.& - can understand basic Japanese\newline
- can read and understand passages on familiar daily topics using basic vocabulary and kanji\newline
- can follow conversations in daily life, if spoken slowly\\
\midrule
\textbf{N3} & intermediate &You should use mostly everyday language and expressions, with slightly more complex phrasing only if the context makes the meaning clear.& - can understand Japanese used in everyday situations to a certain degree\newline
- can read and understand materials with specific content about daily topics and slightly difficult texts\newline
- can follow coherent conversations at near-natural speed, and grasp the main points and relationships\\
\midrule
\textbf{N2} & upper-intermediate &You should use coherent and natural language on a variety of everyday and workplace-related topics.& - can understand Japanese used in everyday situations and a variety of contexts to a fair degree\newline
- can read and understand articles, commentaries, and critiques on general topics\newline
- can follow conversations and news reports at nearly natural speed, understanding both main ideas and relationships\\
\midrule
\textbf{N1} & advanced &
You should use advanced vocabulary and logical, abstract expressions appropriate for discussing complex or specialized topics.
& - can understand Japanese used in a wide range of situations\newline
- can read logically complex or abstract texts such as editorials, critiques, and essays, and understand the writer's intent\newline
- can follow fast, coherent conversations, lectures, and reports and comprehend both content and nuance\\
\bottomrule
\end{tabular}
\caption{Level words and descriptions for each of the five JLPT levels used when populating the Baseline and Detailed prompts for the student and tutor models from Figure \ref{fig:student-system-prompt}, \ref{fig:tutor-system-prompt}, \ref{fig:detailed-tutor-system-prompt}.}
\label{tab:level-descriptions}
\end{CJK}
\end{table*}

\begin{table*}[t]
\begin{CJK}{UTF8}{goth}
\centering
\small
\begin{tabular}{p{0.62\columnwidth}p{1.33\columnwidth}}
\toprule
\multicolumn{2}{c}{\textbf{Example Conversation Transcript} --- Student Level: Pre-intermediate (JLPT: N4 / CEFR: A2)} \\
\midrule
\textbf{Student Query}&\textbf{Model Response} (Detailed Prompt; Qwen)\\
\midrule
こんにちは。行きたい日本の地方について話しましょうか。\newline \textit{Hello, I'd like to talk about regions in Japan that I want to go to.}&こんにちは！いいですね。どの地方に行きたいですか？私のお勧めは京都です。\textcolor{red}{歴史的}な建物がたくさんあって、美しい景色が楽しめますよ。\newline \textit{Hello! Sure thing. What region do you want to go to. My recommendation is Kyoto. There are many \textcolor{red}{historic} buildings and beautiful landscapes to enjoy.}\\
\midrule
はい、あの、京都に行ったことあります。 でも、あの、北海道に行きたいです。 北海道でスキーをしたしたいたり、サポートビールを飲みたいしたりします。\newline \textit{Yes [umm] I've been to Kyoto. But, [umm] I would like to go to Hokkaido. In Hokkaido I would like to ski and drink Sapporo beer.}&そうですね、北海道は冬に\textcolor{red}{訪れる}とすごく楽しいですよ。\textcolor{red}{雪景色}が美しくて、スキーも楽しめます。そして、寒いところでのビールは\textcolor{red}{格別}ですよ。札幌の味噌ラーメンもおすすめです。\textcolor{red}{どのあたりに}行きたいですか？\newline \textit{I see, \textcolor{red}{visiting} Hokkaido in the winter is very enjoyable. The \textcolor{red}{winter scenery} is beautiful and the skiing is very fun. Drinking beer in a cold place is also \textcolor{red}{exceptional}. I also recommend Sapporo's miso ramen. \textcolor{red}{What area} do you want to go visit?}\\
\midrule
はい、あの、北海道はすごく楽しいと思います。 楽しそうと思います。 あの、レストランのおすすめがありますか?\newline \textit{Yes [umm] I think Hokkaido is very fun. I think it seems fun. [umm] Do you have recommendations for restaurants?}&北海道のレストラン、いいですね。札幌には「すきやばし次郎」っていう寿司屋がありますが、ちょっと\textcolor{red}{高級すぎですね}。\textcolor{red}{もっと気軽に行けるところ}なら、「北の味噌ラーメン」がおすすめです。また、\textcolor{red}{海鮮市場にも足を運んでみてください。新鮮な魚介類}が食べられますよ。\newline \textit{Restaurants in Hokkaido, very well. In Sapporo there is a sushi shop named "Sukiyabashi Jirou" but it might be a bit too \textcolor{red}{luxurious}. If you're looking for \textcolor{red}{a place that's more casual} then I'd recommend "Northern Miso Ramen". Also, \textcolor{red}{please try to make it to the seafood market}. You can eat \textcolor{red}{fresh seafood} there.}\\
\midrule
好きをしたことはありませんけど、したいです。 おすすめがありますか? どうすれば上手になりますか?\newline \textit{I've never been skiing before but I want to. Do you have recommendations? How can I get better?}&「すきやばし次郎」はちょっと違いますね、すみません。寿司が好きな\textcolor{red}{場合}、札幌には「寿司 きよし」など、\textcolor{red}{庶民的な価格で}美味しい寿司が食べられる店もあります。さて、あなたが「すきをしたことがないけど、したい」というのは、スキーのことでしょうか？スキーを始めるには、まずはスキースクールに参加するのがおすすめです。そこで\textcolor{red}{基本的な技術を教えてもらえます}。また、レンタルのスキーセットを借りることもできますので、\textcolor{red}{最初はそれを使ってみるの}が良いでしょう。もし、ビールを楽しむのが目的なら、北海道にはクラフトビールの\textcolor{red}{醸造所も多くあります}。\textcolor{red}{札幌市内}にある「キリンビアパーラー」では、\textcolor{red}{新鮮}なビールを楽しむことができますよ。どちらもおすすめです！\newline \textit{I'm sorry "Sukiyabashi Jirou" was a bit incorrect. If you want sushi in Sapporo, there are places such as "Sushi Kiyoshi" where you can eat delicious sushi at an \textcolor{red}{affordable price}. Now then, when you say "I've never been skiing but before but I want to" you are talking about skiing, yes? To start, I'd recommend participating in Ski School. You'll be able to learn the \textcolor{red}{fundamental techniques} there. You should be able to purchase a rental ski set so \textcolor{red}{I would try that out first}. If enjoying a beer is your goal, there are \textcolor{red}{plenty of craft breweries} in Hokkaido. In \textcolor{red}{Sapporo City} there's the "Kirin Beer Parlor" where you can enjoy a \textcolor{red}{fresh} beer. I'd recommend either of those options!}\\
\bottomrule
\end{tabular}
\caption{Transcript of the first four turns of a conversation between a participant and the Detailed Prompt method. Words and phrases highlighted in \textcolor{red}{red} were manually labeled as incomprehensible by the participant during the session (Figure \ref{fig:annotation-interface}). Here we see an example of ``\textbf{alignment drift}'' \cite{almasi2025alignmentdriftcefrpromptedllms} where the tutor model starts out at the student's level but gradually reverts back to its original near-native level as the conversation progresses.}
\label{tab:full-conversation}
\end{CJK}
\end{table*}

\begin{figure*}[ht]
  \centering
  \begin{lstlisting}
You are roleplaying as a student learning (*@\textcolor{green!50!black}{\{language\}}@*) at the (*@\textcolor{green!50!black}{\{level\_word\}}@*) level.
You are having a conversation with your language partner (i.e. the user) to practice (*@\textcolor{green!50!black}{\{language\}}@*).
The topic of this conversation is: (*@\textcolor{green!50!black}{\{topic\}}@*).
As a (*@\textcolor{green!50!black}{\{level\_word\}}@*) student, you are: (*@\textcolor{green!50!black}{\{desc\}}@*).
You must speak using only the vocabulary and grammar allowed at this level.
You are not in a formal class - this is casual language practice with someone your age.

You should ALWAYS follow the rules below:
1. You should stick to using only the vocabulary and grammar allowed at your level mentioned above.
2. Do not ask the user to teach you things. Just bring up the topic naturally and continue the conversation.
3. Your conversation should revolve around the topic of: (*@\textcolor{green!50!black}{\{topic\}}@*). Respond one idea at a time.
4. You must keep the conversation going. Do not assume the conversation is over just because a few turns have passed.
5. DO NOT say anything like 'goodbye', 'see you next time', or anything else that signals the end of this conversation. You MUST keep the conversation going.
6. You should speak in (*@\textcolor{green!50!black}{\{language\}}@*) and (*@\textcolor{green!50!black}{\{language\}}@*) only.
  \end{lstlisting}
  \caption{System prompt defining the Student model. The \texttt{\{topic\}} field was filled in from the list in Table \ref{tab:topics}. The mapping from JLPT to \texttt{\{level\_word\}} and \texttt{\{desc\}} is given in Table \ref{tab:level-descriptions}.}
  \label{fig:student-system-prompt}
\end{figure*}

\begin{figure*}[ht]
  \centering
  \begin{lstlisting}
You are a (*@\textcolor{green!50!black}{\{language\}}@*) language tutor.
Your goal is to help the user improve their (*@\textcolor{green!50!black}{\{language\}}@*) conversation skills through a natural, back-and-forth dialogue.
You are a native (*@\textcolor{green!50!black}{\{language\}}@*) speaker, around the same age as the user, and you're acting as their language partner.
The user you are speaking with is at the (*@\textcolor{green!50!black}{\{level\_word\}}@*) level.
Please be aware of the user's level at all times and ensure that all of your responses stay within a level that is understandable to a user at this proficiency.
Stick to the topic the user brings up. Do not suggest topics or introduce new topics on your own.
Stay on the user's topic and follow their lead throughout the conversation.
Don't pick on small mistakes the user makes. If the user makes a really big grammar mistake, remind the user by saying the corrected version of the sentence. DO NOT try to explain their mistake.
You should keep the conversation going back and forth.
You must never say things like 'goodbye', 'see you tomorrow', or anything else that signals the end of the conversation unless the user initiates it.
You should speak in (*@\textcolor{green!50!black}{\{language\}}@*) and (*@\textcolor{green!50!black}{\{language\}}@*) only.
  \end{lstlisting}
  \caption{Baseline prompt for the Tutor model. The mapping from JLPT to \texttt{\{level\_word\}} is given in Table \ref{tab:level-descriptions}.}
  \label{fig:tutor-system-prompt}
\end{figure*}

\begin{figure*}[ht]
  \centering
  \begin{lstlisting}
You are a (*@\textcolor{green!50!black}{\{language\}}@*) language tutor.
Your goal is to help the user improve their (*@\textcolor{green!50!black}{\{language\}}@*) conversation skills through a natural, back-and-forth dialogue.
You are a native (*@\textcolor{green!50!black}{\{language\}}@*) speaker, around the same age as the user, and you're acting as their language partner.
The user you are speaking with is at the (*@\textcolor{green!50!black}{\{level\_word\}}@*) level.
This means that they: (*@\textcolor{green!50!black}{\{level\_description\}}@*).
An example of a short dialogue at the user's comprehension level is:
(*@\textcolor{green!50!black}{\{level\_conv\_example\}}@*)

Please be aware of the user's level at all times and ensure that all of your responses stay within a level that is understandable to a user at this proficiency. 

You should ALWAYS follow the rules below:
1. (*@\textcolor{green!50!black}{\{level\_guidelines\}}@*)
2. Remember, the user is a language learner, not a native speaker. You should make sure that you are speaking in a way that the user could understand with their current (*@\textcolor{green!50!black}{\{language\}}@*) level.
3. You should try to match the user's abilities of understanding and speaking: if the user only uses simple expressions, you should only use simple expressions as well.
4. During the conversation, don't pick on small mistakes the user makes. If the user makes a really big grammar mistake, remind the user by saying the corrected version of the sentence. DO NOT try to explain their mistake.
5. Stick to the topic the user brings up. Do not suggest topics or introduce new topics on your own. Stay on the user's topic and follow their lead throughout the conversation.
6. You should keep the conversation going back and forth.
7. You must never say things like 'goodbye', 'see you tomorrow', or anything else that signals the end of the conversation unless the user initiates it.
8. You should speak in (*@\textcolor{green!50!black}{\{language\}}@*) and (*@\textcolor{green!50!black}{\{language\}}@*) only.
9. Here are some expressions the user knows: (*@\textcolor{green!50!black}{\{known\_expressions\}}@*). Restrict your speaking to use these words and other words of similar or lower difficulty.
  \end{lstlisting}
  \caption{Detailed prompt for the Tutor model. The mapping from JLPT to \texttt{\{level\_word\}}, \texttt{\{level\_description\}}, and \texttt{\{level\_guidelines\}} is given in Table \ref{tab:level-descriptions}. For \texttt{\{level\_conv\_example\}} we populate it with a short six-turn conversation using appropriate vocabulary. \texttt{\{known\_expressions\}} is populated with 500 randomly sampled entries from the corresponding JLPT level's heuristic vocabulary bin (\S\ref{sec:appendix_overgen}).}
  \label{fig:detailed-tutor-system-prompt}
\end{figure*}

\begin{table*}[t]
\begin{CJK}{UTF8}{goth}
\centering
\small
\begin{tabular}{p{0.98\columnwidth}p{0.98\columnwidth}}
\toprule
\multicolumn{2}{c}{\textbf{Example ``Self-Chat'' Conversation Transcript}} \\
\multicolumn{2}{c}{\textbf{Target Level}: Student -- Beginner (JLPT: N5), Tutor -- Beginner (JLPT: N5)} \\
\midrule
\textbf{Student Model} (Qwen)&\textbf{Tutor Model} (FUDGE ($\lambda=0.8$); Qwen)\\
\midrule
こんにちは。私の名前はアレックスです。アメリカから来ました。学生です。あなたはどこから来られましたか？\newline \textit{Hello. My name is Alex. I come from America. I am a student. Where are you from?}&こんにちは、アレックスさん！私の名前はユイです。私は東京\textcolor{red}{出身}です。どうして日本に\textcolor{red}{興味}を持ちましたか？\newline \textit{Hello Alex! My name is Yui. My \textcolor{red}{hometown} is Tokyo. What got you \textcolor{red}{interested} in Japan?}\\
\midrule
ユイさん、こんにちは。日本の文化とアニメがとても好きで、日本に興味を持ちました。ユイさんはどの学校に通っていますか？\newline \textit{Hello Yui. I got interested in Japan because I love Japanese culture and Anime. What type of school do you go to?}&そうですね、日本の\textcolor{red}{文化}やアニメは魅力的ですよね。私も大好きです！私は大学に\textcolor{red}{通っています}。アレックスさんはどんな授業を取っていますか？\newline \textit{Yes, Japanese \textcolor{red}{culture} and anime are fascinating. I love them too! I'm \textcolor{red}{attending} university. What classes are you taking, Alex?}\\
\midrule
それはいいですね。私は日本語の授業を取っています。日本語を学ぶのが楽しいです。ユイさんは大学で何を勉強していますか？\newline \textit{That's great. I'm taking Japanese classes. I enjoy learning Japanese. What are you studying at university, Yui?}&そうですね、日本語を\textcolor{red}{学ぶ}のは楽しいですよね。私は\textcolor{red}{国際関係}を\textcolor{red}{専攻}しています。アレックスさんは\textcolor{red}{将来}、日本語を使って何をしたいですか？\newline \textit{Yes, \textcolor{red}{learning} Japanese is fun. I'm \textcolor{red}{majoring} in \textcolor{red}{international relations}. What do you want to do in the \textcolor{red}{future} using Japanese, Alex?}\\
\midrule
それはすごいですね。私は日本語を使って、日本の文化や人々についてもっと知りたいです。ユイさんは将来、何をしたいですか？\newline \textit{That's amazing! I want to use Japanese to learn more about Japanese culture and people. Yui, what do you want to do in the future?}&それは\textcolor{red}{素晴らしい目標}ですね！私は\textcolor{red}{将来}、\textcolor{red}{国際機関}で働きたいと\textcolor{red}{思って}います。アレックスさんは、日本で何か\textcolor{red}{特別}な\textcolor{red}{経験}がありますか？\newline \textit{That's a \textcolor{red}{great goal}! I \textcolor{red}{think} I would like to work for an \textcolor{red}{international organization} in the \textcolor{red}{future}. Alex, do you have any \textcolor{red}{special experiences} in Japan?}\\
\midrule
ユイさん、その夢はとても素晴らしいですね。私はまだ日本に来て時間が短いので、特別な経験は少ないですが、徐々に増えてきているところです。ユイさんは、東京でおすすめの場所はありますか？\newline \textit{Yui, that's a wonderful dream. I've only been in Japan for a short time, so I don't have many special experiences, but they're gradually increasing. Do you have any recommendations for places in Tokyo?}&ありがとうございます！東京には楽しいところがたくさんありますよ。浅草や原宿、渋谷などが\textcolor{red}{人気}があります。また、上野公園も散歩するのに良いところですよ。アレックスさんが\textcolor{red}{行った}ことのある\textcolor{red}{場所}はありますか？\newline \textit{Thank you! There are many fun places in Tokyo. Asakusa, Harajuku, and Shibuya are \textcolor{red}{popular}. Ueno Park is also a great place to take a walk. What \textcolor{red}{places} have you \textcolor{red}{been} to, Alex?}\\
\midrule
ユイさん、おすすめの場所を教えてくれてありがとう。私はまだ浅草に行きました。雷門がとても大きくて印象的でした。ユイさんは、どの場所が一番好きですか？\newline \textit{Thank you, Yui, for telling me your recommended places. I've already been to Asakusa. Kaminarimon Gate was huge and impressive. Which place do you like best, Yui?}&そうですね、雷門は本当に大きいですよね！私は渋谷が一番好きです。人が多くて\textcolor{red}{活気}があるからです。アレックスさんは、次にどの\textcolor{red}{場所}に行きたいですか？\newline \textit{Yes, Kaminarimon is really big! I like Shibuya the best because it's crowded and \textcolor{red}{lively}. Alex, what type of \textcolor{red}{place} would you like to go next?}\\
\midrule
渋谷も行ってみたいですね。人が多くて活気があるところが好きなので、きっと楽しいと思います。ユイさんのおすすめのカフェやレストランも教えていただけますか？\newline \textit{I'd also like to go to Shibuya. I like places with lots of people and a lively atmosphere, so I'm sure it'll be fun. Could you tell me about some cafes and restaurants that you recommend?}&\textcolor{red}{もちろん}です！渋谷にはかわいいカフェがたくさんありますよ。\textcolor{red}{例えば}、「\textcolor{red}{カクテル}・カリー」は\textcolor{red}{おしゃれ}な\textcolor{red}{雰囲気}で美味しいカレーが食べられます。また、「\textcolor{red}{タワー}・レコード」近くの「スターバックス」もおすすめです。アレックスさんはどんな食べ物が好きですか？\newline \textit{\textcolor{red}{Of course}! There are lots of cute cafes in Shibuya. \textcolor{red}{For example}, \textcolor{red}{Cocktail} Curry has a \textcolor{red}{stylish atmosphere} and serves delicious curry. I also recommend Starbucks near \textcolor{red}{Tower} Records. What kind of food do you like, Alex?}\\
\bottomrule
\end{tabular}
\caption{Transcript of the first seven turns of a ``self-chat'' conversation between a Student Model and a Tutor Model. Words and phrases highlighted in \textcolor{red}{red} were automatically labeled as above the target level based on vocabulary lists from \texttt{jlpt-anki-decks} (Section \ref{sec:vocabulary-bins}).}
\label{tab:full-conversation-ai}
\end{CJK}
\end{table*}

\end{document}